\documentclass[a4paper,fleqn]{cas-dc}
\usepackage[authoryear]{natbib}

\usepackage{algorithmic}
\usepackage{algorithm}
\usepackage{tabularx}
\usepackage{textcomp}
\usepackage{verbatim}
\usepackage{pifont}
\usepackage{xurl}
\newcommand{\cmark}{\ding{51}}
\newcommand{\xmark}{\ding{55}}

\begin{document}
\let\WriteBookmarks\relax
\def\floatpagepagefraction{1}
\def\textpagefraction{.001}
\shorttitle{EviDep: Uncertainty-Aware Multimodal Depression Estimation}
\shortauthors{Fangyuan Liu et~al.}

\title[mode = title]{EviDep: Uncertainty-Aware Multimodal Depression Estimation via Disentangled Evidential Learning}

\author[1]{Fangyuan Liu}
\ead{liufangyuan@mail.ustc.edu.cn}
\author[1]{Sirui Zhao}
\cormark[1]
\ead{siruit@ustc.edu.cn}
\author[2]{Yangsong Zhang}
\ead{zhangysacademy@gmail.com}
\author[3]{Jinyang Huang}
\ead{hjy@hfut.edu.cn}
\author[1]{Feng-Qi Cui}
\ead{fengqi_cui@mail.ustc.edu.cn}
\author[4]{Bin Luo}
\ead{luobinsky@foxmail.com}
\author[1]{Tong Xu}
\ead{tongxu@ustc.edu.cn}
\author[1]{Enhong Chen}
\cormark[1]
\ead{cheneh@ustc.edu.cn}

\affiliation[1]{organization={School of Computer Science and Technology, University of Science and Technology of China},
                city={Hefei}, country={China}}
\affiliation[2]{organization={School of Computer Science and Technology, Laboratory for Brain Science and Artificial Intelligence, Southwest University of Science and Technology},
                city={Mianyang}, country={China}}
\affiliation[3]{organization={School of Computer Science and Information Engineering, Hefei University of Technology},
                city={Hefei}, country={China}}
\affiliation[4]{organization={Department of Psychiatry, The First Affiliated Hospital of University of Science and Technology of China, Division of Life Sciences and Medicine, University of Science and Technology of China},
                city={Hefei}, country={China}}
\cortext[cor1]{Corresponding authors.}

\begin{abstract}
Audio--visual recordings provide complementary cues for estimating depression severity, but their informativeness varies across time and modalities. Point predictions alone do not express the uncertainty associated with these estimates. We present EviDep, a multimodal evidential regression framework that integrates multi-scale temporal modeling and shared--private representation learning for uncertainty-aware depression estimation. Frequency-aware Feature Extraction decomposes behavioral feature sequences into multiple frequency bands and refines them with scale-specific experts. Disentangled Evidential Learning encourages the disentanglement of cross-modal shared and modality-specific information in the refined features. Multi-branch Evidential Regression maps the resulting shared and private representations to three Normal-Inverse-Gamma (NIG) outputs and uses evidence-weighted aggregation to estimate depression severity and quantify aleatoric and epistemic uncertainty. Experiments on AVEC 2013, AVEC 2014, DAIC-WOZ, and E-DAIC show competitive prediction accuracy, with ablation studies supporting the contributions of frequency-aware refinement and shared--private disentanglement. Further analyses show that estimated epistemic uncertainty helps identify higher-error predictions, while both uncertainty estimates generally increase under controlled feature degradation.
\end{abstract}



\maketitle

\section{Introduction}
Depression affects how people feel, speak, and interact, making observable behavior a valuable source of information for severity assessment. Its substantial burden on individuals and healthcare systems~\citep{WHO2017depression} has motivated research into automated analysis of speech and facial behavior~\citep{GUOHOU2020103349,heDeepLearningDepression2022}. These signals offer complementary views of an individual's behavior and can support quantitative assessment alongside self-reports and clinical interviews.

Behavioral cues for depression vary across time and modalities. Slowly evolving patterns coexist with brief changes in expression or speech, while silence, occlusion, and recording artifacts can obscure relevant information. A model may therefore produce similar severity estimates from recordings that differ substantially in the quality and consistency of their supporting evidence. For automated depression assessment, the estimated score alone does not indicate how much confidence should be placed in an individual prediction. An inaccurate but apparently confident estimate may mislead subsequent assessment, particularly when behavioral evidence is incomplete or ambiguous. Quantifying predictive uncertainty can help identify estimates that warrant closer review or additional information, making it an important complement to severity prediction.
\begin{figure}[pos=t]
\centering
\includegraphics[width=\linewidth]{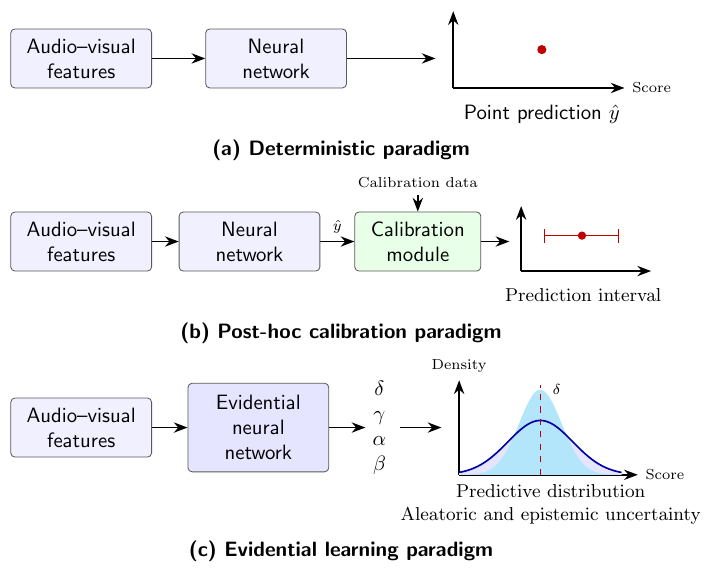}
\caption{Comparison of how depression estimation paradigms represent predictions and uncertainty. (a) Deterministic regression provides a severity estimate alone. (b) Post-hoc calibration supplements the estimate with a prediction interval. (c) EviDep learns NIG distribution parameters to estimate severity and quantify aleatoric and epistemic uncertainty jointly.}
\label{fig:fig1}
\end{figure}

Despite this need, several existing multimodal depression estimation methods~\citep{MAFF,li2025audio,10851289} focus primarily on representation learning for deterministic regression. Although advances in temporal modeling and cross-modal interaction have improved prediction accuracy, these methods produce scalar severity estimates $\hat{y}$ without explicitly quantifying predictive uncertainty (Fig.~\ref{fig:fig1}(a)). Recent studies have addressed uncertainty through conformal prediction, constructing calibrated prediction intervals with coverage guarantees under exchangeability~\citep{10887078,liFairUncertaintyQuantification2025} (Fig.~\ref{fig:fig1}(b)). Other work has leveraged uncertainty for task reweighting to improve predictive performance and fairness~\citep{pmlr-v259-cheong25a}. Complementing these efforts, we investigate how the temporal and cross-modal structure of behavioral cues can be incorporated into representations learned jointly for severity estimation and evidential uncertainty quantification.

To this end, we propose EviDep, a multimodal evidential regression framework that jointly estimates depression severity and quantifies predictive uncertainty (Fig.~\ref{fig:fig1}(c)). Behavioral cues span multiple temporal scales, from sustained patterns to brief changes in speech and facial expression. To capture this variation, Frequency-aware Feature Extraction (FFE) decomposes feature sequences into wavelet components and refines them with scale-specific experts. A Transformer models broader temporal context in the low-frequency component, while convolutional experts capture local variations in high-frequency components. The refined components are then reconstructed to integrate information across temporal scales within each modality.

Beyond temporal variation, audio and visual features contain both shared information and modality-specific cues. Disentangled Evidential Learning (DEL) therefore learns shared and private representations, encouraging cross-modal consistency in the shared features and separation between the shared and private components while using reconstruction to encourage information retention. Multi-branch Evidential Regression (MER) then assigns separate Normal--Inverse--Gamma (NIG) heads to the shared, audio-private, and visual-private representations, allowing each branch to estimate severity and its associated uncertainty from a different source of information. Their outputs are aggregated according to the learned evidence while accounting for disagreement among branch predictions, yielding the final severity estimate and predictive uncertainty.

Our main contributions are summarized as follows:
\begin{enumerate}
    \item We propose EviDep, a multimodal evidential regression framework that integrates audio--visual representation learning with depression severity estimation and predictive uncertainty quantification.

    \item We develop FFE and DEL to capture multi-scale behavioral variations and encourage the separation of shared and modality-specific information. MER models these representations with separate NIG heads and uses their learned evidence strengths to guide prediction fusion.

    \item Experiments on four depression benchmarks demonstrate competitive severity estimation performance, with ablation studies supporting the effectiveness of the proposed components. Further analyses show that higher predictive uncertainty is associated with larger estimation errors and that uncertainty increases under controlled feature degradation.
\end{enumerate}
\section{Related Work}
\subsection{Multimodal Depression Estimation}
Multimodal depression estimation integrates audio and visual behavioral cues for severity prediction~\citep{heDeepLearningDepression2022}. Existing methods capture temporal context through unified spatiotemporal networks~\citep{MAFF} and self-attention architectures~\citep{fan2024transformer}. Cross-modal interaction and interpretability have also received attention: \citet{10.1145/3746027.3762062} introduced a collaborative transformer to model bidirectional interactions between modalities, while \citet{10943944} developed a visualization-based mechanism to interpret modality contributions. Alongside these architectural developments, other studies have investigated the refinement and organization of behavioral features. In speech-based depression estimation, \citet{yang2023filterbanks} combined learnable time-domain filterbanks with multi-scale spectral attention to identify depression-related frequency bands, while \citet{yang2026disnet} explored interpretable representations through learnable frequency-domain filterbanks and hierarchical feature extraction. \citet{wang2025automatic} addressed feature quality by incorporating noise elimination into long-term behavior modeling. For cross-modal representation learning, \citet{li2025audio} separated modality-shared and modality-specific information. Related work in audiovisual speech modeling~\citep{sadok2024mdvae} used a dynamical variational autoencoder to distinguish shared from modality-specific dynamics while separately modeling sequence-level static information. Building on these advances, EviDep integrates multi-scale temporal modeling and shared--private representation learning with branch-wise evidential regression for joint severity estimation and uncertainty quantification.
\subsection{Uncertainty-Aware Learning}
Existing studies have addressed uncertainty in both training annotations and model predictions. To handle noisy supervision, \citet{he2022noisyannotations} proposed a self-adaptation network that combines sample weighting and relabeling for facial-image-based depression estimation, while \citet{10251764} represented ambiguity in PHQ-8 scores through label distribution learning. Beyond uncertainty in supervision, \citet{nilsen2022delta} developed an efficient Delta-method approximation to quantify epistemic uncertainty in deep classifiers. \citet{10887078} introduced conformal prediction with marginal coverage guarantees under exchangeability and an extension targeting approximate conditional coverage. \citet{liFairUncertaintyQuantification2025} addressed demographic disparities in interval coverage through group-conditional conformal prediction and fairness-aware optimization. Complementing these approaches, EviDep couples audio--visual representation learning with evidential regression to jointly estimate depression severity and quantify the associated aleatoric and epistemic uncertainty.

\subsection{Deep Evidential Learning}
Deep evidential learning models uncertainty through distributions over likelihood parameters. For fMRI-based diagnostic classification, \citet{ye2025bpen} proposed the Brain Posterior Evidential Network (BPEN), which uses normalizing flows to estimate class-conditional latent densities and update a Dirichlet prior over class probabilities. For continuous prediction, deep evidential regression (DER)~\citep{NEURIPS2020_aab08546} learns the parameters of a Normal--Inverse--Gamma (NIG) distribution over the mean and variance of a Gaussian likelihood, yielding point predictions and model-based estimates of aleatoric and epistemic uncertainty in a single forward pass. \citet{Wu_Shi_Dong_Zheng_Wei_2024} investigated evidence contraction in DER and proposed a non-saturating uncertainty regularizer to alleviate it. Extending evidential regression to multimodal data, \citet{ma2021trustworthy} introduced the Mixture of Normal--Inverse--Gamma distributions (MoNIG) framework, which aggregates evidential predictions from individual modalities and a feature-fused pseudo-modality. EviDep builds on NIG-based evidential regression and aggregation, while organizing its prediction branches around shared, audio-private, and visual-private representations learned from frequency-refined audio--visual features.
\section{Methodology}
\begin{figure*}[pos=t]
\centering
\includegraphics[width=\linewidth]{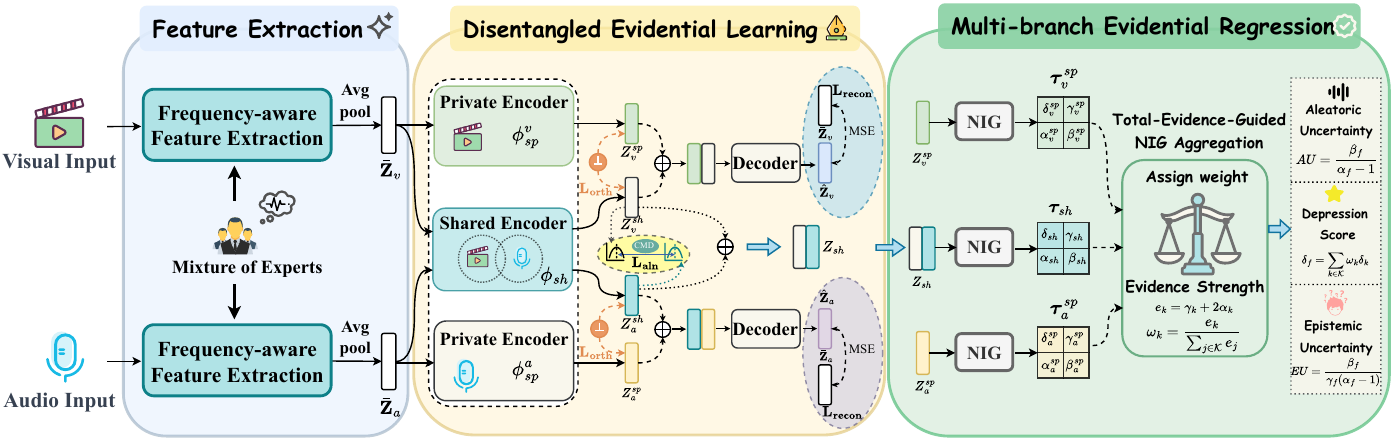}
\caption{Overall architecture of EviDep. FFE extracts multi-scale temporal features, and DEL learns shared and modality-specific representations. MER maps these representations to three NIG outputs and aggregates them to estimate depression severity and quantify predictive uncertainty.}
\label{fig:framework}
\end{figure*}
\subsection{Framework Overview}
As illustrated in Fig.~\ref{fig:framework}, EviDep consists of three stages: Frequency-aware Feature Extraction (FFE), Disentangled Evidential Learning (DEL), and Multi-branch Evidential Regression (MER). Given audio and visual feature sequences, FFE models temporal dependencies and refines frequency components to capture behavioral variations at different temporal scales. DEL then learns shared and private representations from the pooled features, using alignment and orthogonality to encourage disentanglement and reconstruction to help preserve information. These representations are passed to MER, where three NIG heads predict distribution parameters from cross-modal shared, audio-private, and visual-private information. Evidence-weighted aggregation fuses their outputs to obtain the final depression severity estimate and its aleatoric and epistemic uncertainties. The framework is jointly optimized with evidential learning and representation regularization objectives.

\subsection{Frequency-aware Feature Extraction}
\subsubsection{Temporal Context Modeling}
For each modality $m \in \{v, a\}$, we first employ a Transformer encoder to model intra-modal temporal dependencies from the input feature sequence $\mathbf{X}_m \in \mathbb{R}^{T \times D_m}$, yielding an intermediate representation $\mathbf{H}_m \in \mathbb{R}^{T \times D}$ for subsequent frequency-aware refinement.
\begin{equation}
    \mathbf{H}_m = \text{TransformerEncoder}_m(\mathbf{X}_m),
\end{equation}
where $T$ denotes the temporal length of the sequence, $D_m$ the input feature dimension of modality $m$, and $D$ the hidden feature dimension.
\subsubsection{Frequency-aware Feature Refinement}

\begin{figure}[pos=t]
\centering
\includegraphics[width=\linewidth]{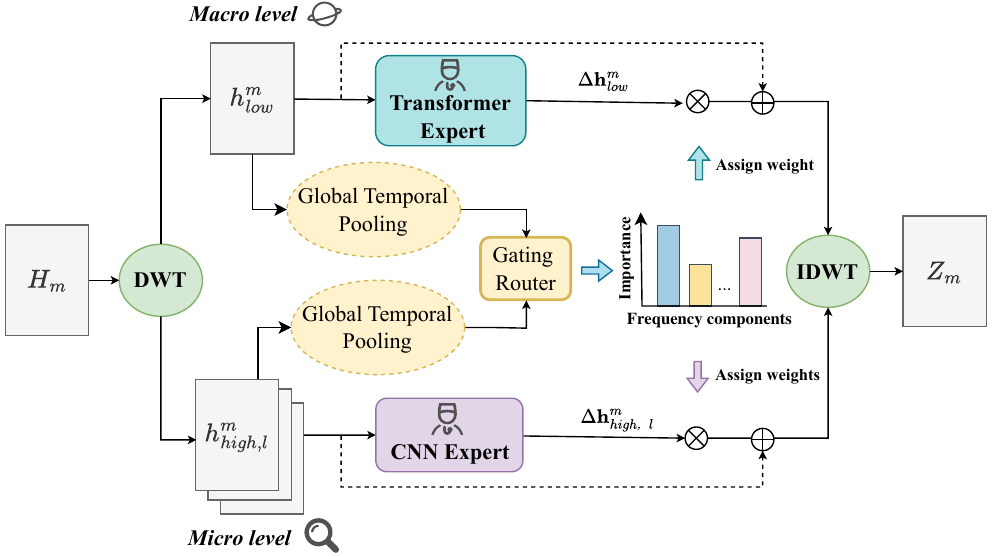}
\caption{Architecture of the Frequency-aware Feature Refinement module, comprising DWT-based decomposition, heterogeneous Transformer-CNN Mixture-of-Experts, adaptive gating, and IDWT-based reconstruction.}
\label{fig_freq}
\end{figure}

To capture variations at different temporal scales, the frequency-aware refinement module decomposes the encoded feature sequences into wavelet components, refines them with scale-specific experts, and reconstructs the enhanced features, as illustrated in Fig.~\ref{fig_freq}.

An $L$-level Discrete Wavelet Transform (DWT) decomposes the temporal features $\mathbf{H}_m$ into a low-frequency approximation and high-frequency detail components. The approximation captures slower variation in the learned feature sequence, while the detail components describe changes at successive temporal resolutions:
\begin{equation}
\{\mathbf{h}_{high,l}^m\}_{l=1}^L, \mathbf{h}_{low}^m = \text{DWT}(\mathbf{H}_m).
\end{equation}

A scale-adaptive Mixture-of-Experts (MoE) architecture assigns a dedicated expert to each component. CNN experts use local receptive fields to refine high-frequency components, while a Transformer expert uses global self-attention to refine the low-frequency component:
\begin{equation}
\small
\Delta \mathbf{h}_{i}^m = 
\begin{cases} 
\text{CNN}_{i}^{m}(\mathbf{h}_{i}^m), & i \in \{high_1, \dots, high_L\} \\
\text{Transformer}^{m}(\mathbf{h}_{i}^m), & i = low
\end{cases}
\end{equation}

We introduce a gating vector $\mathbf{g}_m \in (0,1)^{L+1}$ to adaptively control the residual updates of the frequency components. Each entry $g_{m,i}$ scales the update produced by the expert assigned to component $i$, determining how much of that update is added to the original component. To compute these weights, we first apply temporal average pooling to each frequency component, obtaining $\bar{\mathbf{h}}_i^m \in \mathbb{R}^{D}$. The pooled features are then concatenated and passed to a modality-specific MLP, so that each gating weight is determined using information from all frequency components:
\begin{equation}
\begin{aligned}
\mathbf{w}_m &= \left[\bar{\mathbf{h}}_{high,1}^m ; \dots ; \bar{\mathbf{h}}_{high,L}^m ; \bar{\mathbf{h}}_{low}^m\right],\\
\mathbf{g}_m &= \sigma\!\left(\text{MLP}_m(\mathbf{w}_m)\right),
\end{aligned}
\end{equation}
where $[\cdot;\cdot]$ denotes feature concatenation, $\mathbf{w}_m \in \mathbb{R}^{(L+1)D}$, and $\sigma$ is the element-wise sigmoid function. The resulting weights are applied through gated residual connections:
\begin{equation}
\tilde{\mathbf{h}}_i^m = \mathbf{h}_i^m + g_{m,i} \cdot \Delta \mathbf{h}_{i}^m, \quad i \in \{high_1, \dots, high_L, low\},
\end{equation}

The gated residual updates retain the original frequency components while incorporating the refinements learned by the experts. The updated components are reconstructed into a temporal feature sequence $\mathbf{Z}_m \in \mathbb{R}^{T \times D}$ using the Inverse Discrete Wavelet Transform (IDWT). We apply an element-wise GELU activation to the reconstructed features before temporal average pooling, obtaining the representation $\bar{\mathbf{Z}}_m \in \mathbb{R}^{D}$ for DEL:
\begin{equation}
\begin{aligned}
\mathbf{Z}_m &= \text{IDWT}\!\left(\{\tilde{\mathbf{h}}_{high,l}^m\}_{l=1}^L, \tilde{\mathbf{h}}_{low}^m\right),\\
\bar{\mathbf{Z}}_m &= \frac{1}{T}\sum_{t=1}^{T}\operatorname{GELU}\!\left(\mathbf{Z}_m(t,\cdot)\right).
\end{aligned}
\end{equation}

\subsection{Disentangled Evidential Learning}
Audio and visual features contain both cross-modal commonality and modality-specific cues that may contribute differently to severity estimation. DEL encourages the disentanglement of shared and modality-specific information, providing a structured basis for the prediction branches in MER. The shared pathway is trained to capture cross-modal commonality, while the private pathways model modality-specific variation. Alignment and orthogonality guide this disentanglement, and reconstruction helps retain information from the refined features. MER models the resulting shared, audio-private, and visual-private representations through distinct evidential prediction branches. It aggregates their predictions according to the learned evidence, allowing branch contributions to vary across recordings.

\subsubsection{Modality-Shared and Modality-Private Representations}
For each modality $m \in \{v, a\}$, we project the globally pooled representation $\bar{\mathbf{Z}}_m \in \mathbb{R}^{D}$ into a cross-modal shared component $\mathbf{z}_{m}^{sh} \in \mathbb{R}^{D}$ and a modality-specific component $\mathbf{z}_{m}^{sp} \in \mathbb{R}^{D}$. Specifically, $\bar{\mathbf{Z}}_m$ is fed into two parallel projection heads with identical architectures (a two-layer MLP with GELU activation) but different parameterization schemes:
\begin{equation}
\mathbf{z}_{m}^{sh} = \phi_{sh}(\bar{\mathbf{Z}}_m), \quad \mathbf{z}_{m}^{sp} = \phi_{sp}^m(\bar{\mathbf{Z}}_m),
\end{equation}
where $\phi_{sh}$ uses the same parameters for both modalities, applying a common mapping to their pooled features. The modality-specific projector $\phi_{sp}^m$ has separate parameters for each modality, allowing its private representation to model distinct variations. The shared and private pathways are trained to capture cross-modal consistency and modality-specific variation through the objectives below.

The two shared components are concatenated to form the input to the shared evidential branch:
\begin{equation}
\mathbf{z}_{sh} = [\mathbf{z}_{v}^{sh} ; \mathbf{z}_{a}^{sh}] \in \mathbb{R}^{2D},
\end{equation}
where $[\cdot;\cdot]$ denotes concatenation. The concatenated shared representation $\mathbf{z}_{sh}$ and the two private representations serve as the three inputs to MER.
\subsubsection{Disentanglement and Alignment Constraints}
We use cross-modal alignment and orthogonality regularization to encourage shared--private disentanglement.

\paragraph{Cross-Modal Consistency} We encourage the shared representations $\mathbf{z}_{v}^{sh}$ and $\mathbf{z}_{a}^{sh}$ to have similar distributions using Central Moment Discrepancy (CMD)~\citep{zellinger2017cmd}:
\begin{equation}
\mathcal{L}_{aln} = \text{CMD}(\mathbf{z}_{v}^{sh}, \mathbf{z}_{a}^{sh}).
\end{equation}

Here, CMD is computed over the shared representations in each training batch, matching their means and element-wise central moments of orders two through five. This objective encourages distributional alignment between the audio and visual shared representations.

\paragraph{Orthogonality Constraint}
To encourage shared--private disentanglement, we penalize directional overlap between the shared and private representations within each modality. The same constraint is applied between the audio-private and visual-private representations:
\begin{equation}
\mathcal{L}_{orth}=
\sum_{m\in\{v,a\}}
\ell_{orth}(\mathbf{z}_{m}^{sh},\mathbf{z}_{m}^{sp})
+\ell_{orth}(\mathbf{z}_{v}^{sp},\mathbf{z}_{a}^{sp}),
\label{eq:aggregated_orth}
\end{equation}
where $\ell_{orth}$ measures the squared cosine similarity between two representations:
\begin{equation}
\ell_{orth}(\mathbf{u},\mathbf{v})=
\left(\frac{\mathbf{u}^{\top}\mathbf{v}}
{\|\mathbf{u}\|_2\|\mathbf{v}\|_2}\right)^2.
\label{eq:pairwise_orth}
\end{equation}

Together, the alignment and orthogonality objectives encourage the disentanglement of shared and modality-specific information, while reconstruction helps retain information needed for subsequent evidential prediction.

\subsubsection{Reconstruction Regularization}
To retain information from the refined features during disentanglement, we reconstruct each pooled feature vector from its shared and private representations. For each modality, a decoder $\mathcal{D}_m$ receives the concatenated components and reconstructs the pooled feature as $\hat{\mathbf{Z}}_m = \mathcal{D}_m([\mathbf{z}_{m}^{sh} ; \mathbf{z}_{m}^{sp}])$.

Each decoder $\mathcal{D}_m$ uses Layer Normalization followed by a two-layer MLP to map the concatenated $2D$-dimensional factors back to a $D$-dimensional feature vector. The process is optimized with a mean-squared reconstruction loss:
\begin{equation}
    \mathcal{L}_{rec} = \sum_{m \in \{v, a\}} \frac{1}{D}\|\bar{\mathbf{Z}}_m - \hat{\mathbf{Z}}_m\|_2^2,
\end{equation}
where $\hat{\mathbf{Z}}_m$ is the reconstruction of the pooled feature $\bar{\mathbf{Z}}_m$.

This objective encourages the shared and private representations to jointly retain the information needed to recover the pooled input.
\subsection{Multi-branch Evidential Regression}
MER maps the shared and private representations learned by DEL to branch-wise NIG parameters and aggregates them for severity estimation and uncertainty quantification.
\subsubsection{Joint Modeling of Aleatoric and Epistemic Uncertainty}
Multimodal depression estimation involves aleatoric uncertainty ($\text{AU}$), associated with variability and ambiguity in the data, and epistemic uncertainty ($\text{EU}$), associated with limited model knowledge. To estimate both within an evidential regression framework~\citep{NEURIPS2020_aab08546}, we model the depression severity score $y$ using a Gaussian likelihood, $y\mid\mu,\sigma^2\sim\mathcal{N}(\mu,\sigma^2)$, and place a Normal-Inverse-Gamma (NIG) conjugate prior over its unknown mean $\mu$ and variance $\sigma^2$:
\begin{equation}
p(\mu, \sigma^2 \mid \boldsymbol{\tau}) = \text{NIG}(\mu, \sigma^2 \mid \delta, \gamma, \alpha, \beta),
\end{equation}
where $\boldsymbol{\tau}=(\delta,\gamma,\alpha,\beta)$ denotes the NIG parameters predicted from the corresponding branch representation.

Here, $\delta$ is the predictive mean and serves as the branch severity estimate. The parameter $\gamma$ controls the conditional variance of $\mu$ through $\mu\mid\sigma^2\sim\mathcal{N}(\delta,\sigma^2/\gamma)$, while $\alpha$ and $\beta$ are the shape and scale parameters of $\sigma^2\sim\operatorname{InvGamma}(\alpha,\beta)$. For $\alpha>1$, the NIG moments provide the following aleatoric and epistemic uncertainty estimates:
\begin{equation}
\text{AU} = \mathbb{E}[\sigma^2] = \frac{\beta}{\alpha-1}, \quad \text{EU} = \text{Var}(\mu) = \frac{\beta}{\gamma(\alpha-1)}.
\end{equation}

Here, AU denotes the expected observation variance, whereas EU quantifies uncertainty in the mean parameter $\mu$. Integrating out $\mu$ and $\sigma^2$ yields a Student-$t$ predictive distribution with location $\delta$, $2\alpha$ degrees of freedom, and squared scale $s^2=\beta(1+\gamma)/(\gamma\alpha)$. By the law of total variance, the predictive variance equals $\text{AU}+\text{EU}$.

\subsubsection{Evidence Parameterization}
Each shared or private representation is mapped to its own evidential parameters. For branch $k \in \{sh, v_{sp}, a_{sp}\}$, a dedicated regression head predicts $\boldsymbol{\tau}_k = (\delta_k, \gamma_k, \alpha_k, \beta_k)$:
\begin{equation}
\boldsymbol{\tau}_k = \text{NIGHead}_k(\mathbf{z}_k).
\end{equation}

\begin{figure}[pos=t]
\centering
\includegraphics[width=\linewidth, height=4cm, keepaspectratio]{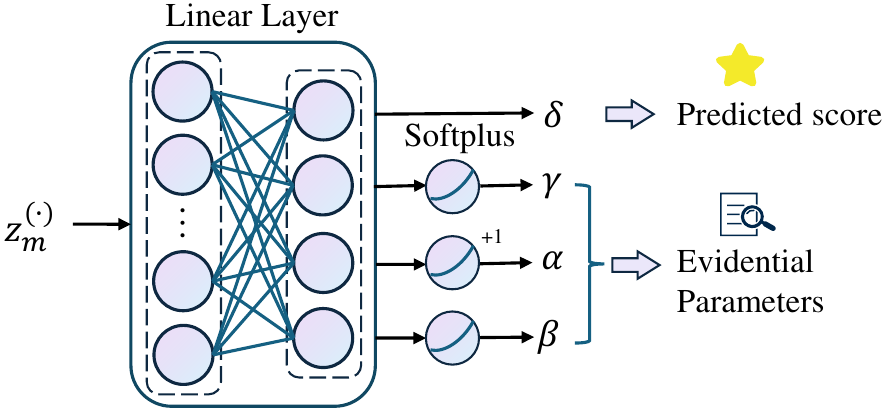}
\caption{Detailed structure of the NIG Regression Head. The latent feature is projected into a 4-D vector via a linear layer, with specific Softplus-based activations to ensure the NIG parameters ($\delta, \gamma, \alpha, \beta$) satisfy their respective domain constraints.}
\label{fig_4}
\end{figure}

As illustrated in Fig.~\ref{fig_4}, each NIG Head linearly projects the input features $\mathbf{z}_k$ into a four-dimensional vector $\mathbf{v} = [v_1, v_2, v_3, v_4]^\top$. A valid NIG distribution requires $\delta\in\mathbb{R}$ and $\gamma,\alpha,\beta>0$. We impose the stronger condition $\alpha>1$ so that the variance estimates above are finite, using the following parameterization:
\begin{equation}
\begin{split}
\delta &= v_1, \\
\gamma &= \operatorname{softplus}(v_2) + \epsilon, \\
\alpha &= \operatorname{softplus}(v_3) + 1, \\
\beta  &= \operatorname{softplus}(v_4) + \epsilon,
\end{split}
\end{equation}
where $\epsilon = 10^{-6}$ is a small constant for numerical stability. This parameterization ensures that each branch separately yields valid NIG parameters.

\subsubsection{Total-Evidence-Guided NIG Aggregation}
\label{sec:evidence_aggregation}
The shared, audio-private, and visual-private branches estimate depression severity from different aspects of the learned representations and may therefore yield different predictions. MER aggregates their NIG outputs to obtain a unified estimate while accounting for both the learned branch evidence and disagreement among predictions. Motivated by multimodal evidential aggregation~\citep{ma2021trustworthy}, we determine the contribution of each branch using an evidence score derived from its NIG parameters. For $k\in\mathcal{K}=\{sh,a_{sp},v_{sp}\}$, with $K=|\mathcal{K}|=3$, the score and its normalized weight are:
\begin{equation}
e_k=\gamma_k+2\alpha_k,
\qquad
\omega_k=\frac{e_k}{\sum_{j\in\mathcal{K}}e_j}.
\label{eq:total_evidence_weight}
\end{equation}

Here, $\gamma_k$ controls the precision of the likelihood mean relative to the observation variance, while $\alpha_k$ governs the shape of the inverse-gamma variance distribution. The factor of two expresses the latter on the degrees-of-freedom scale of the induced Student-$t$ distribution. Using their sum as the aggregation score allows both parameters to influence the relative contribution of each branch. The fused severity estimate is:
\begin{equation}
\delta_f=\sum_{k\in\mathcal{K}}\omega_k\delta_k.
\label{eq:evidence_consensus_mean}
\end{equation}
Equivalently, this estimate minimizes the evidence-weighted squared deviations from the branch predictions:
\begin{equation}
\delta_f
=\arg\min_{z\in\mathbb{R}}
\sum_{k\in\mathcal{K}}e_k(z-\delta_k)^2.
\label{eq:consensus_objective}
\end{equation}

To carry information about branch disagreement into the fused distribution, we aggregate the remaining NIG parameters as:
\begin{equation}
\begin{aligned}
\gamma_f &= \sum_{k\in\mathcal{K}}\gamma_k,\\
\alpha_f &= \sum_{k\in\mathcal{K}}\alpha_k+\frac{K-1}{2},\\
\beta_f &= \sum_{k\in\mathcal{K}}\beta_k
+\frac{1}{2}\sum_{k\in\mathcal{K}}
\gamma_k(\delta_k-\delta_f)^2.
\end{aligned}
\label{eq:evidence_nig_aggregation}
\end{equation}

The fused scale $\beta_f$ contains two contributions: the sum of the branch scale parameters and a correction for deviations of the branch means from $\delta_f$. The correction is weighted by $\gamma_k$ and vanishes when all branch means agree. For fixed $\gamma_f$ and $\alpha_f$, this nonnegative correction increases the fused predictive variance when the branch means differ. Branch disagreement is therefore reflected in the uncertainty accompanying the consensus estimate.

The aggregation is invariant to branch ordering and yields valid NIG parameters with $\gamma_f>0$, $\alpha_f>1$, and $\beta_f>0$. The final severity estimate is $\delta_f$, with aleatoric and epistemic uncertainty obtained from the fused NIG moments.

\subsubsection{Evidential Regression Loss}
For a branch or fused output $\boldsymbol{\tau} = (\delta, \gamma, \alpha, \beta)$, we minimize the negative log-likelihood (NLL) of the induced Student-$t$ predictive distribution~\citep{NEURIPS2020_aab08546}:
\begin{equation}
\begin{split}
    \mathcal{L}_{nll} &= \frac{1}{2}\ln\left(\frac{\pi}{\gamma}\right) - \alpha\ln(\Omega) + C(\alpha) \\
    &\quad + \left(\alpha + \frac{1}{2}\right)\ln\left(\gamma(y - \delta)^2 + \Omega\right),
\end{split}
\end{equation}
where $y$ is the ground-truth score, $C(\alpha) = \ln\frac{\Gamma(\alpha)}{\Gamma(\alpha + 0.5)}$, $\Gamma(\cdot)$ denotes the Gamma function, and $\Omega = 2\beta(1 + \gamma)$ is a shorthand for the NLL expression.

We adopt the evidence regularizer introduced by \citet{NEURIPS2020_aab08546} for deep evidential regression. This regularizer penalizes large values of $\gamma$ and $\alpha$ in proportion to the absolute prediction residual:
\begin{equation}
\mathcal{L}_{reg} = |y - \delta| \cdot (2\gamma + \alpha).
\end{equation}

The evidential loss combines likelihood fitting and evidence regularization:
\begin{equation}
\label{eq:evid_loss}
\mathcal{L}_{\text{EDL}} = \mathcal{L}_{nll} + \lambda_{r} \mathcal{L}_{reg},
\end{equation}
where $\lambda_{r}$ balances likelihood fitting and the incorrect-evidence penalty. 
\subsection{Total Optimization Objective}
The total optimization objective jointly optimizes representation disentanglement and evidential prediction through structural regularization and supervision of the branch and fused outputs.

First, the structural loss combines the aggregated orthogonality objective in Eq.~\eqref{eq:aggregated_orth}, cross-modal alignment, and reconstruction:
\begin{equation}
\mathcal{L}_{\text{struct}} = 
\lambda_{\text{orth}}\mathcal{L}_{\text{orth}} +
\lambda_{\text{aln}} \mathcal{L}_{\text{aln}} + 
\lambda_{\text{rec}} \mathcal{L}_{\text{rec}}.
\end{equation}

Second, let $\mathcal{K}_{aux} = \{sh, v_{sp}, a_{sp}\}$ denote the three NIG heads. The evidential objective combines supervision of the aggregated NIG with branch-wise auxiliary supervision:
\begin{equation}
\mathcal{L}_{\text{evid}} = \mathcal{L}_{\text{EDL}}^{(f)} + \lambda_{\text{aux}} \sum_{k \in \mathcal{K}_{aux}} \mathcal{L}_{\text{EDL}}^{(k)},
\end{equation}
where $\lambda_{\text{aux}}$ controls deep supervision.

To stabilize early training~\citep{11217233}, we gradually decrease the weight of the MSE loss while increasing that of the evidential objective. The two objectives are weighted by $1-\eta_t$ and $\eta_t$, respectively, where $\eta_t=\min(1,t/T_{\mathrm{warm}})$. The total training objective is:
\begin{equation}
\mathcal{L}_{\text{total}}^{(t)} = \mathcal{L}_{\text{struct}} + (1 - \eta_t) \mathcal{L}_{\text{MSE}}(\delta_f, y) + \eta_t \mathcal{L}_{\text{evid}},
\end{equation}
where $t$ denotes the current epoch, and $T_{\mathrm{warm}}$ is the warm-up duration.

\section{Experiments}
Our experiments address three questions. How accurately does EviDep estimate depression severity across different recording settings? What do its predictive distributions reveal beyond point error? Which representation and prediction components contribute to the observed behavior? We address these questions through benchmark comparisons, uncertainty diagnostics, component ablations, and controlled feature degradation.
\subsection{Datasets}
\subsubsection{AVEC 2013}
The AVEC 2013 dataset~\cite{valstar2013avec} is a subset of the Audio-Visual Depressive Language Corpus, containing 150 video clips from 82 subjects aged between 18 and 63. The recordings primarily capture participants performing specified tasks on a computer, such as reading predefined text or providing improvised responses. Each video is labeled with a depression severity score ranging from 0 to 63 based on the BDI-II scale. 
\subsubsection{AVEC 2014}
The AVEC 2014 dataset~\citep{valstar2014avec} consists of 150 primary audio-visual clips. Each of these clips is partitioned into two distinct sub-clips, independently capturing the Northwind and Freeform tasks. In contrast to the 2013 corpus, the sub-clips in this edition are considerably shorter, with durations ranging from 6 seconds to just over 4 minutes. Consistent with the previous dataset, each primary clip is assigned a single BDI-II score, yielding a continuous depression severity label ranging from 0 to 63.
\subsubsection{DAIC-WOZ}
The DAIC-WOZ dataset~\citep{gratch-etal-2014-distress} is a widely adopted benchmark for automated depression detection. It consists of multimodal clinical interview data, including audio, video, and transcribed text, collected through interactions between 189 participants and a virtual interviewer named Ellie. The corpus spans roughly 50 hours of recordings and provides various low- and high-level feature descriptors, as well as raw speech signals. Depression severity is quantified using PHQ-8 scores on a 0-24 scale.
\subsubsection{E-DAIC}
The Extended Distress Analysis Interview Corpus (E-DAIC)~\citep{ringeval2019avec} expands upon the DAIC-WOZ dataset and was utilized in the AVEC 2019 challenge, comprising a total of 275 clinical interview recordings. The dataset is partitioned into 163 for training, 56 for validation, and 56 for testing. For the multimodal input, E-DAIC provides raw audio, textual transcriptions, and precomputed facial features. Correspondingly, depression severity is labeled using PHQ-8 scores as the ground truth for the regression task.
\subsection{Experimental Setups and Evaluation Metrics}
\subsubsection{Implementation Details}
All experiments use the official dataset splits. For AVEC 2013/2014, visual features are extracted using a ResNet-50 encoder pre-trained on WebFace~\citep{zhu2021webface260m} and fine-tuned only on the training split using video-level labels, yielding 128-dimensional embeddings. For DAIC-WOZ and E-DAIC, we follow~\citet{li2025audio} to derive 49-dimensional visual descriptors comprising head pose, eye gaze, and facial action units. For the 50-dimensional audio representation, Silero VAD~\citep{silero2024vad} identifies speech-active regions. We extract 23 eGeMAPS low-level descriptors using openSMILE and 13 MFCCs with their 13 first-order deltas using librosa. These acoustic descriptors are temporally aligned with the visual features and concatenated with a frame index. 

During training, we randomly sample one 1,000-frame segment per AVEC 2014 clip and five uniformly spaced segments per recording for AVEC 2013, DAIC-WOZ, and E-DAIC. Segments shorter than the required length are zero-padded at the end. Given the pronounced severity imbalance in DAIC-WOZ and E-DAIC, high-severity training samples are oversampled.

Each labeled recording is treated as an evaluation sample. At test time, ten equidistant segments are extracted from each sample. Let $\boldsymbol{\tau}_{is}=(\delta_{is},\gamma_{is},\alpha_{is},\beta_{is})$ denote the fused NIG parameters for segment $s$ of sample $i$. We obtain the sample-level NIG parameters by element-wise averaging over its $S_i=10$ segments:
\begin{equation}
\bar{\boldsymbol{\tau}}_i
=\frac{1}{S_i}\sum_{s=1}^{S_i}\boldsymbol{\tau}_{is}
=(\bar\delta_i,\bar\gamma_i,\bar\alpha_i,\bar\beta_i).
\label{eq:subject_aggregation}
\end{equation}

The final severity estimate is $\hat y_i=\bar\delta_i$. The summarized NIG parameters induce a Student-$t$ predictive distribution with location $\bar\delta_i$, $2\bar\alpha_i$ degrees of freedom, and squared scale $s_i^2=\bar\beta_i(1+\bar\gamma_i)/(\bar\gamma_i\bar\alpha_i)$. We compute the sample-level uncertainty estimates and the total predictive variance of this Student-$t$ distribution from the same parameters:
\begin{equation}
\begin{aligned}
\mathrm{AU}_i &= \frac{\bar\beta_i}{\bar\alpha_i-1},\\
\mathrm{EU}_i &= \frac{\bar\beta_i}{\bar\gamma_i(\bar\alpha_i-1)}, \\
\operatorname{Var}(y_i\mid\bar{\boldsymbol{\tau}}_i)
&=\mathrm{AU}_i+\mathrm{EU}_i.
\end{aligned}
\label{eq:subject_au_eu}
\end{equation}

EviDep is implemented in PyTorch and trained on an NVIDIA GeForce RTX 4090 GPU. The model is trained for up to 100 epochs using Adam with a learning rate of $1\times10^{-4}$ and a batch size of 4. We employ the Haar wavelet basis with a DWT decomposition level of $L=3$ and set the feature dimension to $D=128$. The loss weights are $\lambda_{\text{aln}}=1.0$, $\lambda_{\text{rec}}=0.001$, $\lambda_{\text{aux}}=0.2$, $\lambda_{\text{orth}}=0.05$, and $\lambda_r=1\times10^{-3}$. The curriculum coefficient $\eta_t$ increases linearly from 0 to 1 over the initial $T_{\mathrm{warm}}=20$ warm-up epochs.
Model selection and early stopping use sample-level MAE on the development split.

\subsubsection{Evaluation Metrics}
To comprehensively assess the performance of depression severity estimation, we use Mean Absolute Error (MAE) and Root Mean Square Error (RMSE) to quantify errors. MAE and RMSE measure the average magnitude of prediction errors, defined as $\text{MAE} = \frac{1}{N} \sum |y_i - \hat{y}_i|$ and $\text{RMSE} = \sqrt{\frac{1}{N} \sum (y_i - \hat{y}_i)^2}$.

To assess how well uncertainty ranks prediction errors, we use the area under the sparsification error (AUSE). Predictions are removed in descending order of epistemic uncertainty $\mathrm{EU}_i$, and the MAE of the retained samples is compared with an oracle ranking that removes predictions in descending order of absolute error:
\begin{equation}
\mathrm{AUSE}
=\int_{0}^{0.8}\!\left[M_{\mathrm{EU}}(r)-M_{\mathrm{oracle}}(r)\right]\,\mathrm{d}r,
\label{eq:ause}
\end{equation}
where $M_{\mathrm{EU}}(r)$ and $M_{\mathrm{oracle}}(r)$ denote retained-sample MAE at rejection fraction $r$ under uncertainty and oracle ranking, respectively. The integral is evaluated using the trapezoidal rule over 50 equally spaced rejection fractions from 0 to 0.8. Lower AUSE indicates closer agreement with the oracle in terms of retained-sample MAE.

\begin{table}[width=\linewidth,pos=t]
\centering
\caption{Comparison with state-of-the-art methods on AVEC 2013 and AVEC 2014 datasets. \textbf{Bold} and \textit{italic} denote best and second-best performance, respectively.}
\label{tab:result_avec}

\small
\setlength{\tabcolsep}{3pt}
\renewcommand{\arraystretch}{1.2}

\begin{tabularx}{\linewidth}{@{} l *{4}{>{\centering\arraybackslash}X} @{}}
\toprule
\multirow{2}{*}{\textbf{Methods}} & 
\multicolumn{2}{c}{\textbf{AVEC 2013}} & 
\multicolumn{2}{c}{\textbf{AVEC 2014}} \\
\cmidrule(lr){2-3} \cmidrule(l){4-5}
 & \textit{MAE}$\downarrow$ & \textit{RMSE}$\downarrow$ & \textit{MAE}$\downarrow$ & \textit{RMSE}$\downarrow$ \\
\midrule
\multicolumn{5}{l}{\textbf{\textit{Audio Modality}}} \\
MAFF~\citep{MAFF}                                 & 7.14 & 9.50 & 7.65 & 9.13 \\
TTFNet~\citep{11017633}                           & 7.08 & 8.93 & 7.13 & 8.96 \\
DCCANet~\citep{zhao2024dense}                     & 6.78 & 8.47 & 6.17 & 7.54 \\
TFCAV~\citep{NIU2021208}                          & 6.26 & 8.32 & 7.49 & 9.25 \\
WavDepressionNet~\citep{niu2023wavdepressionnet}  & 6.14 & 8.20 & 6.60 & 8.61 \\
\midrule

\multicolumn{5}{l}{\textbf{\textit{Visual Modality}}} \\
AD-DCNN~\citep{AD-DCNN}                           & 7.58 & 9.82 & 7.47 & 9.55 \\
MAFF~\citep{MAFF}                                 & 7.32 & 8.97 & 6.43 & 8.60 \\
SAN~\citep{he2022noisyannotations}    & 7.02 & 9.37 & 7.28 & 9.31 \\
DJ-LDML~\citep{DJ-LDML}                           & 6.63 & 8.37 & 6.59 & 8.30 \\
DepressNet~\citep{DepressNet}                     & 6.20 & 8.28 & 6.21 & 8.39 \\
MTDAN~\citep{MTDAN}                               & 6.14 & 8.08 & 6.35 & 7.93 \\
MSN~\citep{MSN}                                   & 5.98 & 7.90 & 5.82 & 7.61 \\
TSTM~\citep{10572478}                             & 5.95 & 7.57 & 5.86 & 7.18 \\
MemRank~\citep{10935667} & 5.82 & 7.78 & 5.77 & 7.69 \\
LDBM~\citep{wang2025automatic}                    & 5.71 & 6.99 & 5.36 & 6.93 \\
Depressformer~\citep{HE2024106490}                & 5.49 & 7.47 & 5.56 & 7.22 \\
MLM-EOE~\citep{lin2025mlm}                        & 5.49 & 6.84 & 5.51 & 7.22 \\
\midrule

\multicolumn{5}{l}{\textbf{\textit{Multimodal (Audio + Visual)}}} \\
FedDAAM~\citep{he2025feddaam}                     & 6.78 & 8.61 & 6.77 & 8.59 \\
TMFE-GFN~\citep{fan2024transformer}               & 6.60 & 9.00 & 7.05 & 9.45 \\
AVA-DepressNet~\citep{AVA-DepressNet}             & 6.23 & 7.99 & 5.32 & 6.83 \\
FDFNet~\citep{li2025audio}                        & 6.22 & 7.58 & 5.21 & \textit{6.49} \\
MAFF~\citep{MAFF}                    & 6.14 & 8.16 & 5.21 & 7.03 \\
Dis2DR~\citep{pan2024disentangled}                & 6.12 & 7.97 & 5.45 & 6.61 \\
MFMamba~\citep{liu2025mfmamba}                    & 6.11 & 7.05 & 5.16 & 6.71 \\
VLDSP-TAP-MFB~\citep{9786615}                     & 5.38 & \textit{6.83} & \textit{5.03} & \textbf{6.16} \\
SIMMA~\citep{10932863} & 5.35 & 6.97 & 5.14 & 6.95 \\
STE-Mamba~\citep{lin2025ste}                      & \textit{5.08} & 6.94 & 5.10 & 6.77 \\ \midrule

\textbf{Ours}                                    & \textbf{4.91} & \textbf{6.54} & \textbf{5.02} & 6.80 \\
\bottomrule
\end{tabularx}
\end{table}
\subsection{Comparison with Existing Methods}
We compare EviDep with existing audio-only, visual-only, and audio--visual methods on four public depression benchmarks. As shown in Tables~\ref{tab:result_avec} and~\ref{tab:result_daic}, EviDep achieves the lowest MAE on AVEC 2013, AVEC 2014, and E-DAIC, and the lowest RMSE on AVEC 2013, DAIC-WOZ, and E-DAIC among the compared methods. These results demonstrate its competitive performance in depression severity estimation across the four benchmarks.

On AVEC 2013, EviDep achieves the best performance in both metrics, with an MAE of 4.91 and an RMSE of 6.54. On AVEC 2014, it achieves the lowest MAE of 5.02, although VLDSP-TAP-MFB~\citep{9786615} obtains a lower RMSE. These results support the effectiveness of EviDep for severity estimation from the task-based audio--visual recordings in the AVEC benchmarks. EviDep also performs well on the interview-based datasets. On DAIC-WOZ, EviDep achieves an RMSE of 4.39, compared with 4.71 for FedDAAM~\citep{he2025feddaam}, while the two methods have similar MAE. On E-DAIC, it achieves the lowest MAE and RMSE of 3.72 and 4.60, respectively, outperforming the compared audio-only, visual-only, and audio--visual methods in both metrics. The results across these recording settings support EviDep as an uncertainty-aware framework that retains strong severity prediction performance. The subsequent analyses examine the informativeness of its uncertainty estimates and the contributions of its individual components.
\begin{table}[width=\linewidth,pos=t]
\centering
\caption{Comparison with state-of-the-art methods on DAIC-WOZ and E-DAIC datasets. \textbf{Bold} and \textit{italic} denote best and second-best performance, respectively.}
\label{tab:result_daic}

\small
\setlength{\tabcolsep}{3pt}
\renewcommand{\arraystretch}{1.2} 
\begin{tabularx}{\linewidth}{@{} l *{4}{>{\centering\arraybackslash}X} @{}}
\toprule
\multirow{2}{*}{\textbf{Methods}} & 
\multicolumn{2}{c}{\textbf{DAIC-WOZ}} & 
\multicolumn{2}{c}{\textbf{E-DAIC}} \\
\cmidrule(lr){2-3} \cmidrule(l){4-5}
 & \textit{MAE}$\downarrow$ & \textit{RMSE}$\downarrow$ & \textit{MAE}$\downarrow$ & \textit{RMSE}$\downarrow$ \\
\midrule

\multicolumn{5}{l}{\textbf{\textit{Audio Modality}}} \\
STFN~\citep{10120969}                             & 5.38 & 6.36 & 5.38 & 6.29 \\
AFN~\citep{AFN}                                   & 5.67 & 6.55 & - & - \\
TTFNet~\citep{11017633}                           & 5.09 & 6.01 & 5.00 & 5.76 \\
MA-DLE~\citep{11488689} & 4.31 & 5.49 & 4.68 & 5.72 \\
AGBiTNet~\citep{LIU2025105359}                    & 4.27 & 5.35 & - & - \\
MFDS-VAN~\citep{PAN2024105704}                    & 4.27 & 5.34 & - & - \\
\midrule

\multicolumn{5}{l}{\textbf{\textit{Visual Modality}}} \\
STE-Mamba~\citep{lin2025ste}                      & 4.89 & 6.28 & 4.58 & 5.99 \\
FaceHOG~\citep{10.1007/978-3-030-34869-4_3}       & 4.64 & 5.98 & - & - \\
HOG-PCA~\citep{HOG-PCA}                           & 4.89 & 6.23 & - & - \\
MLM-EOE~\citep{lin2025mlm}                        & - & - & 5.05 & 6.07 \\
LDBM~\citep{wang2025automatic}                    & - & - & 5.13 & 6.28 \\
\midrule

\multicolumn{5}{l}{\textbf{\textit{Multimodal (Audio + Visual)}}} \\
STE-Mamba~\citep{lin2025ste}                      & 5.18 & 6.24 & 5.05 & 6.21 \\
Dis2DR~\citep{pan2024disentangled}                & 4.69 & 5.49 & 4.32 & 5.35 \\
AVA-DepressNet~\citep{AVA-DepressNet}             & 4.62 & 5.78 & - & - \\
SIMMA~\citep{10932863} & 4.79 & 6.81 & 5.31 & 6.37 \\
FDFNet~\citep{li2025audio}                        & 4.25 & 5.34 & 4.41 & 5.10 \\
FAU-GF~\citep{FU2025129106}                       & - & - & \textit{3.77} & \textit{4.95} \\
CDLD~\citep{11153053} & 3.85 & 5.68 & - & - \\
FedDAAM~\citep{he2025feddaam}                     & \textbf{3.68} & \textit{4.71} & - & - \\ \midrule

\textbf{Ours}                                    & \textit{3.73} & \textbf{4.39} & \textbf{3.72} & \textbf{4.60} \\
\bottomrule
\end{tabularx}
\end{table}
\subsection{Uncertainty Quantification Analysis}
We assess predictive uncertainty through its association with error and its expression in individual predictive distributions. Uncertainty-based grouping examines whether higher-uncertainty samples have larger average errors, while rejection analysis evaluates whether uncertainty can help identify predictions to set aside. Individual cases then illustrate how the severity estimate and uncertainty decomposition describe the same recording.
\subsubsection{Correlation Between Uncertainty and Prediction Error}
To assess whether the estimated uncertainty provides information about prediction difficulty, we examine its association with absolute prediction error on AVEC 2014. We use the sample-level epistemic variance $\mathrm{EU}_i$ in Eq.~\eqref{eq:subject_au_eu} as the uncertainty score for both quantile binning and prediction rejection. Quantile binning compares the average error across groups ordered by uncertainty, while uncertainty-based prediction rejection evaluates how the error of the retained samples changes as predictions with higher uncertainty are progressively removed.

\begin{figure}[pos=t]
\centering
\includegraphics[width=\linewidth]{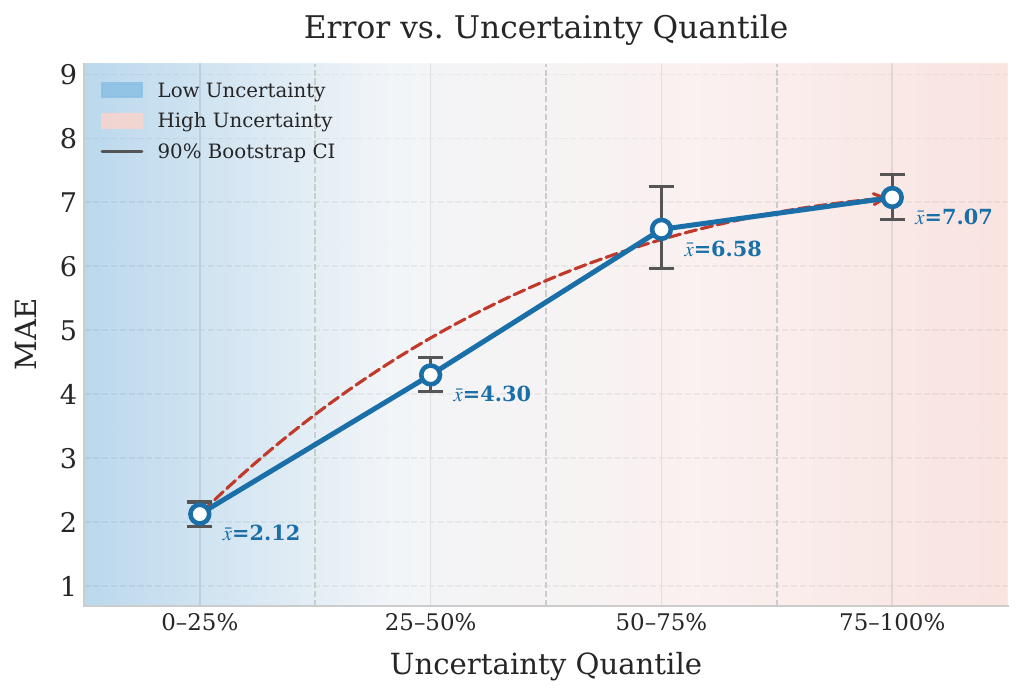}
\caption{MAE and 90\% bootstrap confidence intervals for four groups defined by epistemic uncertainty quartiles on the AVEC 2014 test set. The red dashed arrow highlights the increase in MAE from the lowest- to the highest-uncertainty group.}
\label{fig:error_vs_uncertainty}
\end{figure}

\begin{figure}[pos=t]
\centering
\includegraphics[width=\linewidth]{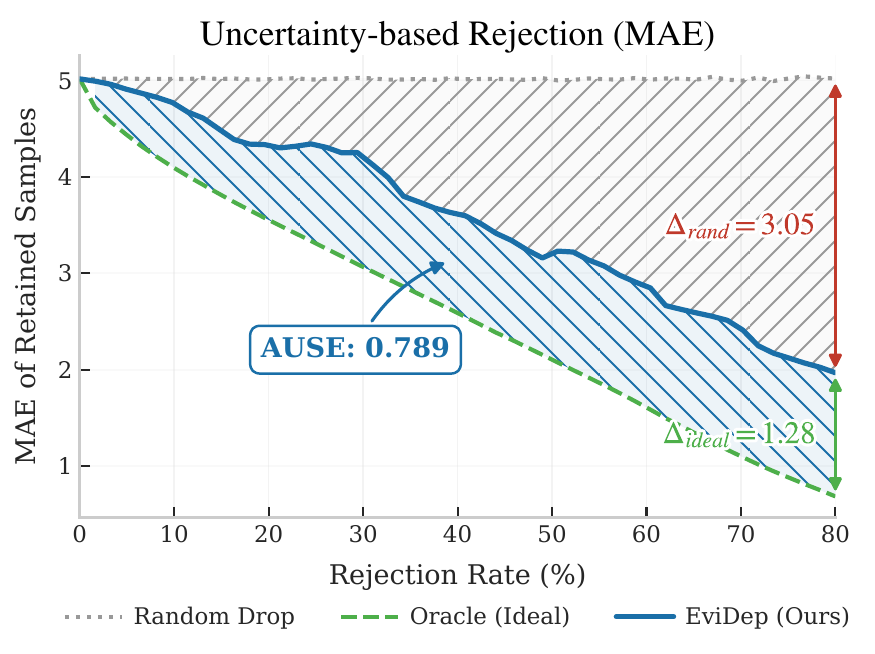}
\caption{Retained-sample MAE under epistemic-uncertainty-based rejection on AVEC 2014, compared with random rejection and an error-ranked oracle.}
\label{fig:sparse}
\end{figure}

We partition the AVEC 2014 test samples into four groups defined by epistemic uncertainty quartiles and compute MAE with 90\% bootstrap confidence intervals in each group. Figure~\ref{fig:error_vs_uncertainty} shows a monotonic increase in average error, from 2.12 in the lowest-uncertainty quartile to 7.07 in the highest. The 4.95-point difference indicates that the uncertainty score distinguishes groups with substantially different average prediction errors.

Figure~\ref{fig:sparse} examines the usefulness of uncertainty ranking through prediction rejection. Predictions are removed in descending order of $\mathrm{EU}_i$, and MAE is computed on the retained subset. Random rejection provides a reference, while the error-ranked oracle orders predictions by their absolute residuals. Using the metric defined in Eq.~\eqref{eq:ause}, EviDep yields an AUSE of 0.789. At 20\% retention, it reduces MAE by 3.05 relative to random rejection, with a remaining gap of 1.28 to the oracle. These results support the use of epistemic uncertainty to identify higher-error predictions.

Together, the increasing quartile errors and the improvement over random rejection support using the estimated uncertainty to rank prediction difficulty on AVEC 2014.
\subsubsection{Qualitative Case Analysis}
\begin{figure}[pos=t]
\centering
\includegraphics[width=\linewidth]{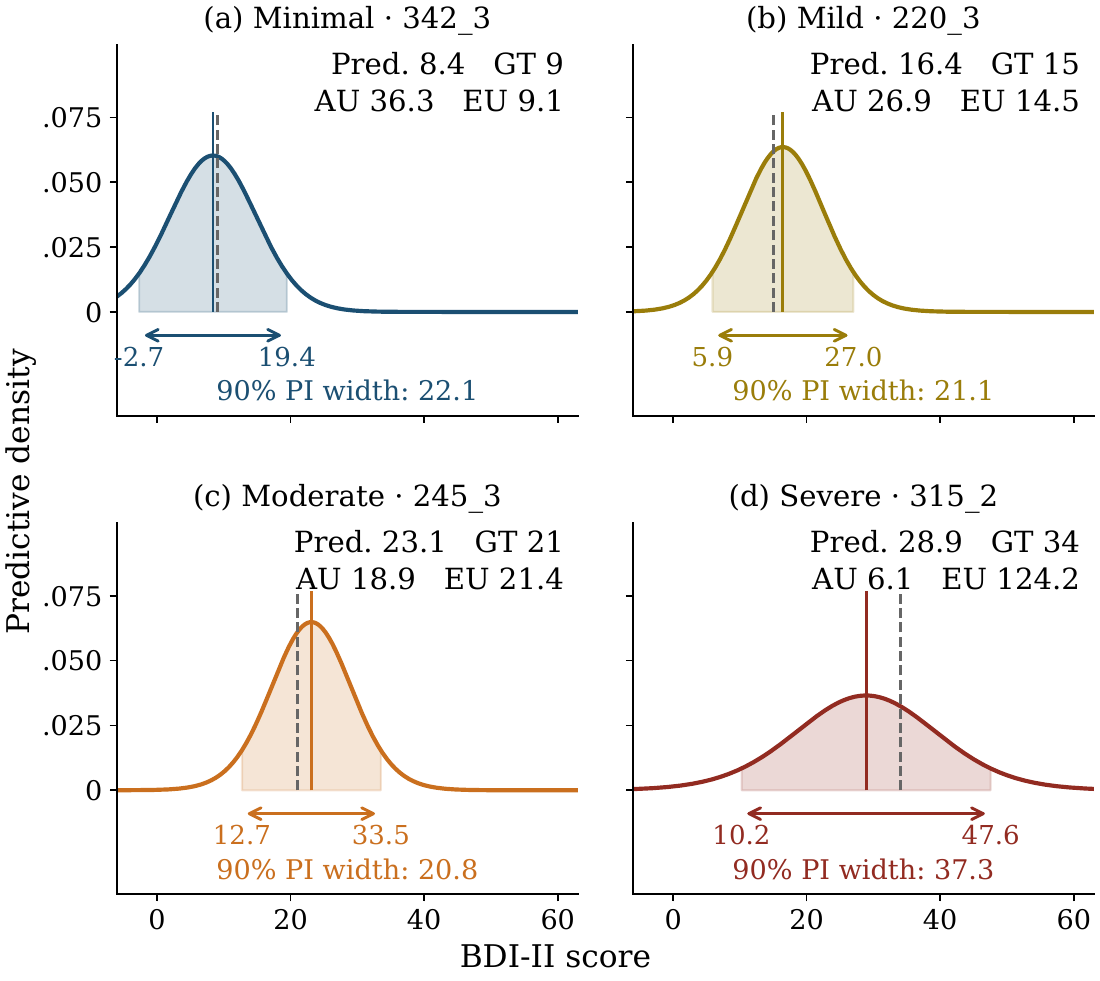}
\caption{Student-$t$ predictive densities derived from the summarized NIG parameters of four representative AVEC 2014 cases. Solid and dashed vertical lines indicate the predicted mean and ground truth. Shading and arrows show the central 90\% prediction interval; panel annotations report its width and the AU and EU computed from the same NIG parameters.}
\label{fig:qualitative_cases}
\end{figure}

Figure~\ref{fig:qualitative_cases} visualizes the Student-$t$ predictive densities and uncertainty estimates for four representative AVEC 2014 samples, using the summarized NIG parameters defined in Eq.~\eqref{eq:subject_aggregation}. The shaded central 90\% prediction interval spans the 5th to 95th percentiles of each density.

For samples 342\_3, 220\_3, and 245\_3, estimated epistemic uncertainty ranges from 9.1 to 21.4. For sample 315\_2, the model predicts 28.9 against a target of 34, with a larger epistemic uncertainty of 124.2 and a more dispersed predictive density. Its estimated aleatoric uncertainty is 6.1, substantially lower than its epistemic uncertainty. The example illustrates why the point estimate alone is incomplete: a severity estimate of 28.9 does not convey how dispersed the corresponding prediction is.

\subsection{Ablation Study}
This section presents component ablations, robustness tests, and parameter sensitivity analyses.

\subsubsection{Impact of Frequency-aware Feature Extraction}
The Frequency-aware Feature Extraction (FFE) module performs temporal context modeling followed by frequency-aware feature refinement. To validate its design, we evaluate the temporal encoder, frequency refinement, expert configuration, and gating mechanism through ablation studies. Table~\ref{tab:ffe_ablation} shows that both temporal modeling and frequency refinement contribute to the performance of FFE. Removing the temporal encoder increases MAE by 0.30 on AVEC 2014 and 0.53 on E-DAIC. Removing frequency refinement also degrades performance, increasing RMSE by 0.39 and 0.59, respectively. 

\begin{table}[width=\linewidth,pos=t]
\centering
\caption{Ablation study of the Frequency-aware Feature Extraction (FFE) module on AVEC 2014 and E-DAIC.}
\label{tab:ffe_ablation}
\small
\renewcommand{\arraystretch}{1.3} 
\setlength{\tabcolsep}{0pt}
\begin{tabular*}{\linewidth}{@{\extracolsep{\fill}}lcccc@{}}
\toprule
\multirow{2}{*}{\textbf{FFE Variant}} 
& \multicolumn{2}{c}{\textbf{AVEC 2014}} & \multicolumn{2}{c}{\textbf{E-DAIC}} \\
\cmidrule(lr){2-3} \cmidrule(lr){4-5}
& \textbf{MAE} $\downarrow$ & \textbf{RMSE} $\downarrow$ 
& \textbf{MAE} $\downarrow$ & \textbf{RMSE} $\downarrow$ \\
\midrule
w/o Temporal Modeling    
& 5.32 & 7.05 & 4.25 & 5.12 \\
w/o Frequency Refinement 
& 5.36 & 7.19 & 4.27 & 5.19 \\
\midrule
w/o MoE
& 5.27 & 6.96 & 4.09 & 4.92 \\
All-CNN Experts          
& 5.22 & 7.01 & 3.91 & 5.11 \\
All-Transformer Experts  
& 5.16 & 6.89 & 4.00 & 4.86 \\
Reversed MoE             
& 5.23 & 7.04 & 4.10 & 5.22 \\
w/o Gating Router        
& 5.11 & 6.98 & 3.93 & 4.73 \\
\midrule
\textbf{Ours}            
& \textbf{5.02} & \textbf{6.80} & \textbf{3.72} & \textbf{4.60} \\
\bottomrule
\end{tabular*}
\end{table}

These results support refining the temporally encoded features at multiple frequency scales, enabling the model to process variations at different temporal resolutions before shared--private representation learning. The w/o MoE variant retains wavelet decomposition and reconstruction but replaces the band-specific experts and gating network with a two-layer MLP shared across frequency bands within each modality, using ungated residual updates. Its higher MAE and RMSE on both datasets support the proposed expert refinement and gating configuration. The choice of experts further affects the effectiveness of frequency refinement. The proposed heterogeneous configuration achieves lower MAE and RMSE than both the all-CNN and all-Transformer alternatives on the two datasets. Reversing the CNN and Transformer assignments increases RMSE by 0.24 on AVEC 2014 and 0.62 on E-DAIC. Residual gating provides an additional improvement within this expert configuration. Without the gating network, RMSE increases from 6.80 to 6.98 on AVEC 2014 and from 4.60 to 4.73 on E-DAIC, accompanied by higher MAE on both datasets. This supports using information from the frequency components to adjust the strength of their expert refinements, allowing FFE to regulate the residual updates before reconstructing the temporal features.

\subsubsection{Validity of Disentangled Evidential Learning}
Part A of Table~\ref{tab:del_ablation} evaluates the objectives that guide shared--private disentanglement and information retention. Removing the full structural loss increases MAE from 5.02 to 5.37 on AVEC 2014 and from 3.72 to 3.93 on E-DAIC. Omitting alignment, orthogonality, or reconstruction individually also increases both error metrics. In particular, removing orthogonality raises MAE to 5.18 and 3.85, respectively. Removing reconstruction raises MAE to 5.17 and 3.86, supporting its use alongside the alignment and orthogonality objectives. These comparisons assess the contribution of the constraints to severity prediction within the proposed framework.

Part B examines how the learned representations contribute to prediction. The variant without the structural loss retains the shared--private pathways. Entangled Evidential applies two NIG heads directly to the pooled audio and visual features without shared--private factorization. Both shared-only and private-only variants have higher errors than the full model, but their relative performance differs across datasets. The shared-only variant performs better on AVEC 2014, whereas the private-only variant performs better on E-DAIC. This pattern supports retaining both sources of information rather than selecting one pathway for all settings. The entangled variant has higher errors than either restricted variant on both datasets. Finally, using a single NIG head with the shared and private representations increases MAE from 5.02 to 5.15 on AVEC 2014 and from 3.72 to 3.91 on E-DAIC. The full configuration therefore benefits from both the learned representations and their use in separate evidential branches.
\begin{table}[width=\linewidth,pos=t]
\centering
\small
\caption{Comprehensive analysis of the Disentangled Evidential Learning (DEL) module on AVEC 2014 and E-DAIC.}
\label{tab:del_ablation}
\renewcommand{\arraystretch}{1.3} 
\setlength{\tabcolsep}{0pt}
\begin{tabular*}{\linewidth}{@{\extracolsep{\fill}}lcccc@{}}
\toprule
\multirow{2}{*}{\textbf{Configuration}} 
& \multicolumn{2}{c}{\textbf{AVEC 2014}} & \multicolumn{2}{c}{\textbf{E-DAIC}} \\
\cmidrule(lr){2-3} \cmidrule(lr){4-5}
& \textbf{MAE} $\downarrow$ & \textbf{RMSE} $\downarrow$ 
& \textbf{MAE} $\downarrow$ & \textbf{RMSE} $\downarrow$ \\
\midrule
\multicolumn{5}{l}{\textit{Part A: Loss Constraints}} \\
w/o $\mathcal{L}_{struct}$ & 5.37 & 7.33 & 3.93 & 4.72 \\
w/o $\mathcal{L}_{aln}$    & 5.19 & 7.01 & 3.86 & 4.80 \\
w/o $\mathcal{L}_{orth}$   & 5.18 & 7.05 & 3.85 & 4.65 \\
w/o $\mathcal{L}_{rec}$    & 5.17 & 7.02 & 3.86 & 4.61 \\
\midrule
\multicolumn{5}{l}{\textit{Part B: Feature Source Contribution}} \\
Entangled Evidential                 & 5.25 & 7.13 & 4.06 & 5.08 \\ 
Shared-Only ($\boldsymbol{\tau}_{sh}$)        & 5.11 & 6.96 & 3.94 & 4.78 \\
Private-Only ($\boldsymbol{\tau}_{v,a}^{sp}$) & 5.20 & 7.08 & 3.89 & 4.64 \\ 
Single NIG Head                               & 5.15 & 6.94 & 3.91 & 4.82 \\
\midrule
\textbf{Ours} & \textbf{5.02} & \textbf{6.80} & \textbf{3.72} & \textbf{4.60} \\
\bottomrule
\end{tabular*}
\end{table}

\begin{figure}[pos=t]
\centering
\includegraphics[width=\linewidth]{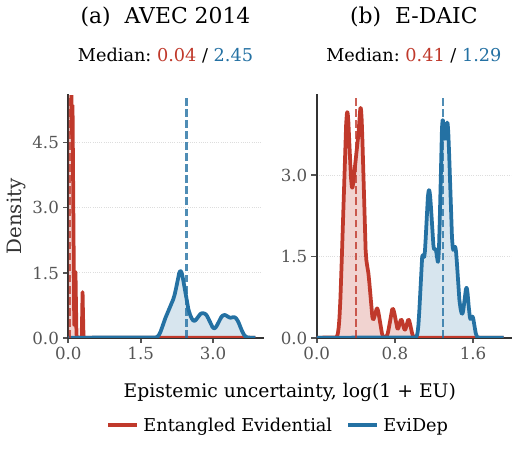}
\caption{Density distributions of predicted epistemic uncertainty for EviDep and the Entangled Evidential baseline in Table~\ref{tab:del_ablation}, on (a) AVEC 2014 and (b) E-DAIC.}
\label{fig:eu}
\end{figure}

Figure~\ref{fig:eu} complements these comparisons by showing the distribution of estimated epistemic uncertainty for EviDep and Entangled Evidential. The entangled baseline concentrates more mass near low uncertainty despite its larger aggregate prediction errors. EviDep produces a different uncertainty distribution alongside lower errors, indicating that the shared--private representation design affects the evidential outputs as well as the predicted scores.

\subsubsection{Evidential Regression and Aggregation}
Table~\ref{tab:evidential_ablation} compares three prediction settings using the same representation backbone architecture: deterministic regression, evidential regression without total-evidence-guided NIG aggregation, and the full MER configuration. The variant without total-evidence aggregation retains the three NIG heads and takes the element-wise arithmetic mean of their output parameters. Its severity estimate is therefore the unweighted mean of the three branch predictions. The full MER configuration instead uses the evidence-weighted mean and disagreement-adjusted scale defined in Section~\ref{sec:evidence_aggregation}. On AVEC 2014, MAE decreases from 5.09 to 5.07 when evidential prediction is introduced and to 5.02 with aggregation; RMSE decreases from 6.92 to 6.88 and then to 6.80. The improvements are larger on E-DAIC, where MAE decreases from 4.34 to 3.96 and then to 3.72. E-DAIC RMSE follows the same ordering, decreasing from 5.23 to 4.69 and then to 4.60. Both evidential prediction and aggregation contribute to the reported accuracy, with different gains across datasets. 

\begin{table}[width=\linewidth,pos=t]
\centering
\caption{Ablation study of evidential regression and total-evidence aggregation on AVEC 2014 and E-DAIC.}
\label{tab:evidential_ablation}
\small
\renewcommand{\arraystretch}{1.2}
\setlength{\tabcolsep}{3pt}

\begin{tabular*}{\linewidth}{@{\extracolsep{\fill}}cccccc@{}}
\toprule
\multirow{2}{*}{\makecell{\textbf{Evidential}\\\textbf{(NIG)}}} 
& \multirow{2}{*}{\makecell{\textbf{Evidence}\\\textbf{Aggregation}}} 
& \multicolumn{2}{c}{\textbf{AVEC 2014}} 
& \multicolumn{2}{c}{\textbf{E-DAIC}} \\
\cmidrule(lr){3-4} \cmidrule(lr){5-6}
& 
& \textbf{MAE} $\downarrow$ & \textbf{RMSE} $\downarrow$
& \textbf{MAE} $\downarrow$ & \textbf{RMSE} $\downarrow$ \\
\midrule
\xmark & \xmark & 5.09 & 6.92 & 4.34 & 5.23 \\
\cmark & \xmark & 5.07 & 6.88 & 3.96 & 4.69 \\
\midrule
\cmark & \cmark & \textbf{5.02} & \textbf{6.80} & \textbf{3.72} & \textbf{4.60} \\
\bottomrule
\end{tabular*}
\end{table}

\subsubsection{Robustness to Feature Degradation}
\begin{figure}[pos=t]
\centering
\includegraphics[width=\linewidth]{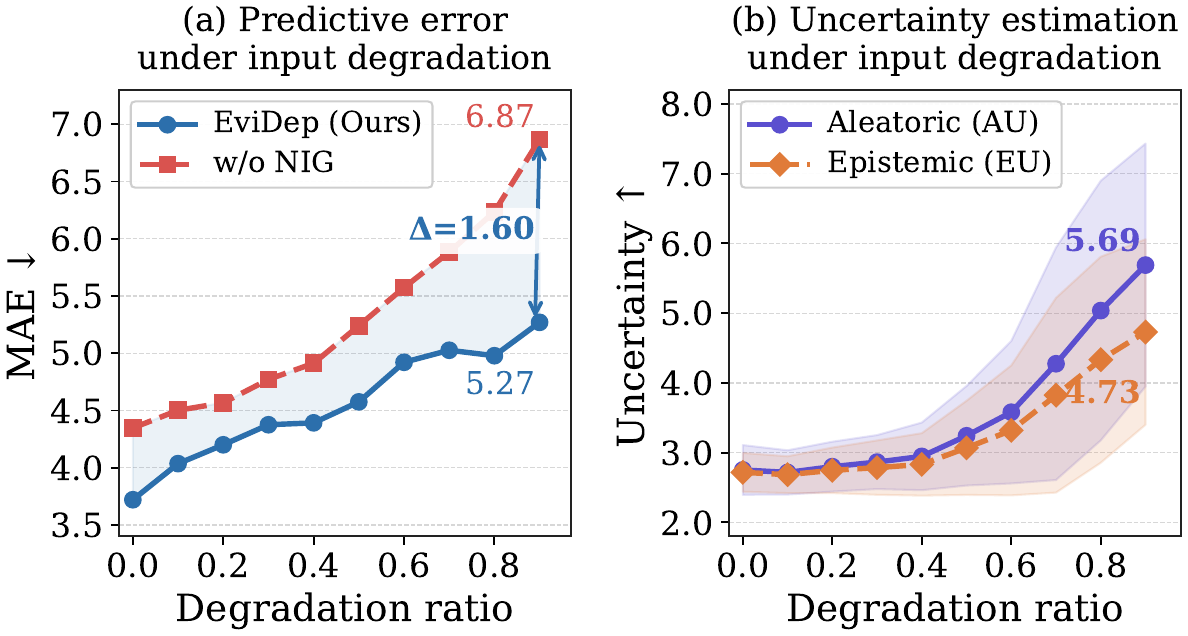}
\caption{Performance and estimated uncertainty under temporal feature degradation on E-DAIC. (a) MAE of the deterministic baseline and EviDep; the shaded area marks their difference. (b) Aleatoric and epistemic uncertainty estimated by EviDep; shaded bands indicate one standard deviation across sample-level uncertainty scores.}
\label{fig:noise_robustness}
\end{figure}

We perturb a contiguous temporal span covering a fraction $r\in[0,0.9]$ of each modality's feature sequence. The same fraction is used for both modalities, with start positions sampled independently. Gaussian noise is added to all feature dimensions within each selected span, scaled by the standard deviation of the unperturbed features. This controlled setting examines how prediction error and uncertainty respond to temporal feature degradation.

As shown in Fig.~\ref{fig:noise_robustness}a, prediction error generally increases for both models as the degradation ratio grows. At a degradation ratio of 0.9, EviDep has an MAE of 5.27, compared with 6.87 for the deterministic baseline, giving a difference of 1.60. This indicates better retention of prediction accuracy under the tested degradation procedure. Both aleatoric and epistemic uncertainty generally increase with stronger degradation (Fig.~\ref{fig:noise_robustness}b). Their rise accompanies the increase in prediction error, indicating that the distributional outputs respond to degraded input features. At the highest degradation ratio, estimated aleatoric and epistemic uncertainty reach 5.69 and 4.73, respectively. Together, the error and uncertainty curves show how the model responds to multimodal feature degradation under this controlled procedure.

\subsubsection{Parameter Sensitivity Analysis}
\begin{figure}[pos=t]
\centering
\includegraphics[width=\linewidth]{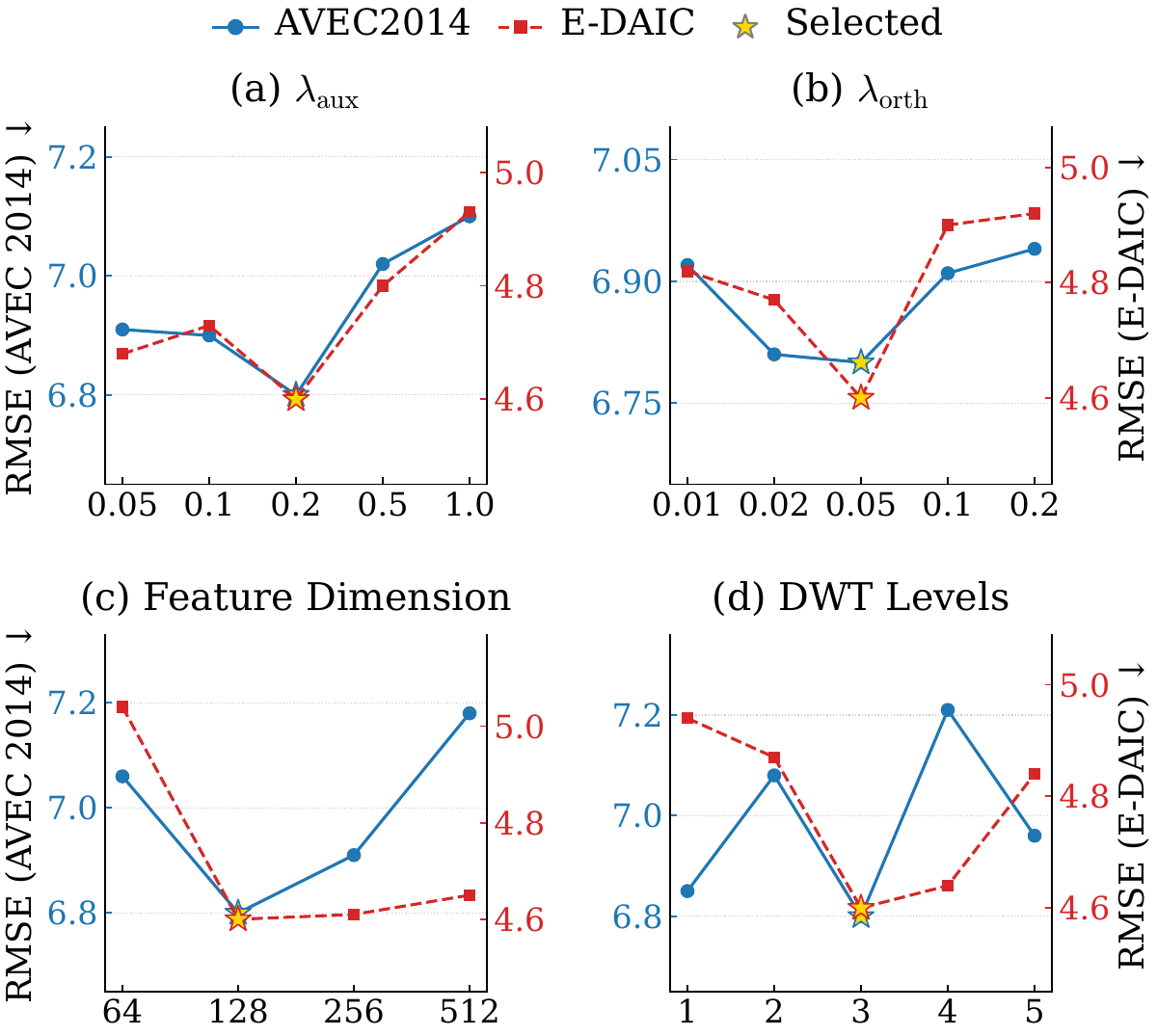}
\caption{Parameter sensitivity analysis on the AVEC 2014 and E-DAIC test sets. We evaluate (a) $\lambda_{aux}$, (b) $\lambda_{orth}$, (c) representation dimension $D$, and (d) DWT levels. Stars ($\star$) mark the default configurations selected on the development splits.}
\label{fig:param_sensitivity}
\end{figure}

Figure~\ref{fig:param_sensitivity} reports test-set performance on AVEC 2014 and E-DAIC for four hyperparameters: auxiliary supervision weight, orthogonality weight, representation dimension, and wavelet depth. The default hyperparameters are selected using the development splits. Each parameter is varied while the remaining settings are held at their defaults.

Within the evaluated grid, $\lambda_{aux}=0.2$, $\lambda_{orth}=0.05$, a 128-dimensional representation, and a three-level wavelet decomposition yield the lowest reported errors. The auxiliary and orthogonality weights both exhibit an intermediate minimum: increasing either weight beyond the selected setting does not further reduce error. Likewise, larger representation dimensions and deeper wavelet decompositions do not consistently improve performance. These results support the selected settings within the evaluated ranges.

\section{Conclusion}
We presented EviDep, a multimodal evidential regression framework for depression severity estimation and uncertainty quantification. FFE captures temporal variation at multiple scales, while DEL encourages the disentanglement of shared and modality-specific information. MER uses these representations for branch-wise NIG prediction and evidence-weighted aggregation to obtain a severity estimate with predictive uncertainty. Experiments across four benchmarks support the predictive utility of these components and show that learned uncertainty contains information about prediction difficulty. Further work should evaluate distributional calibration and generalization across recording conditions. Together, these findings position structured multimodal evidential learning as a direction for estimating not only depression severity, but also the uncertainty accompanying each prediction.
\section*{Acknowledgements}
This work was supported by the National Natural Science Foundation of China under Grant Nos. 62406264 and 61727809.
\section*{CRediT authorship contribution statement}
\textbf{Fangyuan Liu:} Conceptualization, Methodology, Software, Validation, Writing -- original draft, Writing -- review \& editing, Visualization.
\textbf{Sirui Zhao:} Conceptualization, Methodology, Supervision, Writing -- review \& editing.
\textbf{Yangsong Zhang:} Methodology, Writing -- review \& editing.
\textbf{Jinyang Huang:} Writing -- review \& editing.
\textbf{Feng-Qi Cui:} Writing -- review \& editing.
\textbf{Bin Luo:} Writing -- review \& editing.
\textbf{Tong Xu:} Writing -- review \& editing.
\textbf{Enhong Chen:} Supervision, Writing -- review \& editing.
\section*{Declaration of generative AI and AI-assisted technologies
  in the manuscript preparation process}
  During the preparation of this work, the author(s) used OpenAI-ChatGPT to improve the language and readability of the text. The author(s) reviewed and edited the content and take full responsibility for the content of the publication.
\bibliographystyle{cas-model2-names}
\bibliography{ref}

@incollection{valstar2013avec,
  author    = {Valstar, Michel and Schuller, Bj{\"o}rn and Smith, Keith and Eyben, Florian and Jiang, Bowen and Bilakhia, Sharvari and Schnieder, Sebastian and Cowie, Roddy and Pantic, Maja},
  title     = {{AVEC} 2013: The Continuous Audio/Visual Emotion and Depression Recognition Challenge},
  booktitle = {Proceedings of the 3rd {ACM} International Workshop on Audio/Visual Emotion Challenge},
  year      = {2013},
  pages     = {3--10}
}

@incollection{valstar2014avec,
  author    = {Valstar, Michel and Schuller, Bj{\"o}rn and Smith, Keith and Almaev, Timur and Eyben, Florian and Krajewski, J{\"u}rgen and Cowie, Roddy and Pantic, Maja},
  title     = {{AVEC} 2014: {3D} Dimensional Affect and Depression Recognition Challenge},
  booktitle = {Proceedings of the 4th International Workshop on Audio/Visual Emotion Challenge},
  year      = {2014},
  pages     = {3--10}
}

@incollection{ringeval2019avec,
  author    = {Ringeval, Fabien and Schuller, Bj{\"o}rn and Valstar, Michel and Cummins, Nicholas and Cowie, Roddy and Tavabi, Leili and Schmitt, Maximilian and Alisamir, Sina and Amiriparian, Shahin and Messner, Eva-Maria and Song, Siyang and Liu, Shuo and Zhao, Ziping and Mallol-Ragolta, Adria and Ren, Zhao and Soleymani, Mohammad and Pantic, Maja},
  title     = {{AVEC} 2019 Workshop and Challenge: State-of-Mind, Detecting Depression with {AI}, and Cross-Cultural Affect Recognition},
  booktitle = {Proceedings of the 9th International Workshop on Audio/Visual Emotion Challenge and Workshop},
  year      = {2019},
  pages     = {3--12},
  doi = {10.1145/3347320.3357688},
  publisher = {Association for Computing Machinery}
}

@incollection{zhu2021webface260m,
  author    = {Zhu, Zheng and Huang, Guan and Deng, Jiankang and Ye, Yun and Huang, Junjie and Chen, Xinze and Zhu, Jiagang and Yang, Tian and Lu, Jiwen and Du, Dalong and Zhou, Jie},
  title     = {{WebFace260M}: A Benchmark Unveiling the Power of Million-Scale Deep Face Recognition},
  booktitle = {Proceedings of the {IEEE}/{CVF} Conference on Computer Vision and Pattern Recognition ({CVPR})},
  year      = {2021},
  pages     = {10492--10502},
  url = {https://arxiv.org/abs/2103.04098}
}

@article{niu2023wavdepressionnet,
  title={Wavdepressionnet: Automatic depression level prediction via raw speech signals},
  author={Niu, Mingyue and Tao, Jianhua and Li, Yongwei and Qin, Yong and Li, Ya},
  journal={IEEE Transactions on Affective Computing},
  volume={15},
  number={1},
  pages={285--296},
  year={2023},
  publisher={IEEE}
}

@article{NIU2021208,
title = {A time-frequency channel attention and vectorization network for automatic depression level prediction},
journal = {Neurocomputing},
volume = {450},
pages = {208-218},
year = {2021},
issn = {0925-2312},
doi = {10.1016/j.neucom.2021.04.056},
url = {https://www.sciencedirect.com/science/article/pii/S0925231221005981},
author = {Mingyue Niu and Bin Liu and Jianhua Tao and Qifei Li}
}

@incollection{zhao2024dense,
  title={Dense Coordinate Channel Attention Network for Depression Level Estimation from Speech},
  author={Zhao, Ziping and Liu, Shizhao and Niu, Mingyue and Wang, Haishuai and Schuller, Bj{\"o}rn W},
  booktitle={International Conference on Pattern Recognition},
  pages={402--413},
  year={2024},
  organization={Springer}
}

@article{HE2024106490,
title = {Depressformer: Leveraging Video {Swin Transformer} and fine-grained local features for depression scale estimation},
journal = {Biomedical Signal Processing and Control},
volume = {96},
pages = {106490},
year = {2024},
issn = {1746-8094},
doi = {10.1016/j.bspc.2024.106490},
url = {https://www.sciencedirect.com/science/article/pii/S1746809424005482},
author = {Lang He and Zheng Li and Prayag Tiwari and Cui Cao and Jize Xue and Feng Zhu and Di Wu}
}

@article{lin2025mlm,
  title={{MLM-EOE}: Automatic Depression Detection via Sentimental Annotation and Multi-Expert Ensemble},
  author={Lin, Zulong and Wang, Yaowei and Zhou, Yujue and Du, Fei and Yang, Yun},
  journal={IEEE Transactions on Affective Computing},
  year={2025},
  publisher={IEEE}
}

@article{wang2025automatic,
  title={Automatic depression recognition with an ensemble of multimodal spatio-temporal routing features},
  author={Wang, Yaowei and Lin, Zulong and Yang, Chengrong and Zhou, Yujue and Yang, Yun},
  journal={IEEE Transactions on Affective Computing},
  year={2025},
  publisher={IEEE}
}

@article{fan2024transformer,
  title={Transformer-based multimodal feature enhancement networks for multimodal depression detection integrating video, audio and remote photoplethysmograph signals},
  author={Fan, Huiting and Zhang, Xingnan and Xu, Yingying and Fang, Jiangxiong and Zhang, Shiqing and Zhao, Xiaoming and Yu, Jun},
  journal={Information Fusion},
  volume={104},
  pages = {102161},
  year={2024},
  publisher={Elsevier}
}

@article{he2025feddaam,
  title={{FedDAAM}: Federated domain adversarial learning with attention mechanism for privacy preserving multimodal depression assessment},
  author={He, Lang and Yang, Weizhao and Zhao, Junnan and Chen, Haifeng and Jiang, Dongmei},
  journal={IEEE Transactions on Circuits and Systems for Video Technology},
  year={2026},
  volume={36},
  number={2},
  pages={2635--2648},
  doi={10.1109/TCSVT.2025.3609776},
  publisher={IEEE}
}

@article{li2025audio,
  title={Audio-visual Feature Disentanglement and Fusion Network for Automatic Depression Severity Prediction},
  author={Li, Shihao and Shao, Zhuhong and Qin, Rongyin and Huang, Yongzhen and Liang, Peipeng and Li, Xiaobai and Jiang, Yinan and Deng, Yanhe and Liu, Tie and Tan, Xiaohui},
  journal={IEEE Transactions on Affective Computing},
  year={2026},
  publisher={IEEE},
  volume = {17},
  number = {1},
  pages = {190--203},
  doi = {10.1109/TAFFC.2025.3611238}
}

@incollection{pan2024disentangled,
  title={Disentangled-multimodal privileged knowledge distillation for depression recognition with incomplete multimodal data},
  author={Pan, Yuchen and Jiang, Junjun and Jiang, Kui and Liu, Xianming},
  booktitle={Proceedings of the 32nd {ACM} international conference on multimedia},
  pages={5712--5721},
  year={2024}
}

@incollection{liu2025mfmamba,
  title={{MFMamba}: A Multimodal Fusion State Space Model for Depression Recognition},
  author={Liu, Jingyi and Shang, Yuanyuan and Yang, Mengyuan and Shao, Zhuhong and Lu, Jiaxi and Liu, Tie},
  booktitle={{ICASSP} 2025-2025 {IEEE} International Conference on Acoustics, Speech and Signal Processing ({ICASSP})},
  pages={1--5},
  year={2025},
  organization={IEEE}
}

@incollection{lin2025ste,
  title={{Ste-mamba}: Automated multimodal depression detection through emotional analysis and spatio-temporal information ensemble},
  author={Lin, Zulong and Wang, Yaowei and Zhou, Yujue and Du, Fei and Yang, Yun},
  booktitle={{ICASSP} 2025-2025 {IEEE} International Conference on Acoustics, Speech and Signal Processing ({ICASSP})},
  pages={1--5},
  year={2025},
  organization={IEEE}
}

@ARTICLE{11017633,
  author={Chen, Xiyuan and Shao, Zhuhong and Jiang, Yinan and Chen, Runsen and Wang, Yunlong and Li, Bicao and Niu, Mingyue and Chen, Hongguang and Hu, Qiang and Wu, Jiasong and Yang, Chunfeng and Shang, Yuanyuan},
  journal={IEEE Journal of Biomedical and Health Informatics}, 
  title={{TTFNet}: Temporal-Frequency Features Fusion Network for Speech Based Automatic Depression Recognition and Assessment},
  year={2025},
  volume={29},
  number={10},
  pages={7536-7548},
  doi={10.1109/JBHI.2025.3574864}}

@ARTICLE{10120969,
  author={Han, Zhuojin and Shang, Yuanyuan and Shao, Zhuhong and Liu, Jingyi and Guo, Guodong and Liu, Tie and Ding, Hui and Hu, Qiang},
  journal={IEEE Transactions on Cognitive and Developmental Systems}, 
  title={Spatial–Temporal Feature Network for Speech-Based Depression Recognition},
  year={2024},
  volume={16},
  number={1},
  pages={308-318},
  doi={10.1109/TCDS.2023.3273614}}

@incollection{10.1007/978-3-030-34869-4_3,
author="Rathi, Swati
and Kaur, Baljeet
and Agrawal, R. K.",
editor="Deka, Bhabesh
and Maji, Pradipta
and Mitra, Sushmita
and Bhattacharyya, Dhruba Kumar
and Bora, Prabin Kumar
and Pal, Sankar Kumar",
title={Enhanced Depression Detection from Facial Cues Using Univariate Feature Selection Techniques},
booktitle={Pattern Recognition and Machine Intelligence},
year="2019",
publisher="Springer International Publishing",
address="Cham",
pages="22--29",
isbn="978-3-030-34869-4"
}

@incollection{gratch-etal-2014-distress,
    title = {The {Distress Analysis Interview Corpus} of human and computer interviews},
    author = "Gratch, Jonathan  and
      Artstein, Ron  and
      Lucas, Gale  and
      Stratou, Giota  and
      Scherer, Stefan  and
      Nazarian, Angela  and
      Wood, Rachel  and
      Boberg, Jill  and
      DeVault, David  and
      Marsella, Stacy  and
      Traum, David  and
      Rizzo, Skip  and
      Morency, Louis-Philippe",
    editor = "Calzolari, Nicoletta  and
      Choukri, Khalid  and
      Declerck, Thierry  and
      Loftsson, Hrafn  and
      Maegaard, Bente  and
      Mariani, Joseph  and
      Moreno, Asuncion  and
      Odijk, Jan  and
      Piperidis, Stelios",
    booktitle = {Proceedings of the Ninth International Conference on Language Resources and Evaluation ({LREC}'14)},
    month = may,
    year = "2014",
    address = "Reykjavik, Iceland",
    publisher = "European Language Resources Association (ELRA)",
    url = "https://aclanthology.org/L14-1421/",
    pages = "3123--3128",
}

@ARTICLE{10887078,
  author={Li, Yonghong and Qu, Shan and Zhou, Xiuzhuang},
  journal={IEEE Transactions on Affective Computing}, 
  title={Conformal Depression Prediction},
  year={2025},
  volume={16},
  number={3},
  pages={1814-1824},
  doi={10.1109/TAFFC.2025.3542023}}

@techreport{WHO2017depression,
  author      = {{World Health Organization}},
  title       = {Depression and other common mental disorders: global health estimates},
  year        = {2017},
  type        = {Technical Report}
}

@article{GUOHOU2020103349,
title = {What reveals about depression level? {The} role of multimodal features at the level of interview questions},
journal = {Information \& Management},
volume = {57},
number = {7},
pages = {103349},
year = {2020},
issn = {0378-7206},
doi = {10.1016/j.im.2020.103349},
url = {https://www.sciencedirect.com/science/article/pii/S0378720620302871},
author = {Shan, Guohou and Zhou, Lina and Zhang, Dongsong},
}

@misc{liFairUncertaintyQuantification2025,
  title = {Fair Uncertainty Quantification for Depression Prediction},
  author = {Li, Yonghong and Zhang, Zheng and Zhou, Xiuzhuang},
  year = 2025,
  month = sep,
  number = {arXiv:2505.04931},
  eprint = {2505.04931},
  primaryclass = {cs},
  publisher = {arXiv},
  doi = {10.48550/arXiv.2505.04931},
  urldate = {2026-01-13},
  archiveprefix = {arXiv},
  langid = {american},
}

@incollection{pmlr-v259-cheong25a,
  title = 	 {{U-Fair}: Uncertainty-based Multimodal Multitask Learning for Fairer Depression Detection},
  author =       {Cheong, Jiaee and Bangar, Aditya and Kalkan, Sinan and Gunes, Hatice},
  booktitle = 	 {Proceedings of the 4th Machine Learning for Health Symposium},
  pages = 	 {203--218},
  year = 	 {2025},
  editor = 	 {Hegselmann, Stefan and Zhou, Helen and Healey, Elizabeth and Chang, Trenton and Ellington, Caleb and Mhasawade, Vishwali and Tonekaboni, Sana and Argaw, Peniel and Zhang, Haoran},
  volume = 	 {259},
  series = 	 {Proceedings of Machine Learning Research},
  
  publisher =    {PMLR},
  url = 	 {https://proceedings.mlr.press/v259/cheong25a.html},
}

@incollection{10.1145/3746027.3762062,
author = {Liu, Fangyuan and Zhao, Sirui and Yin, Kang and Xu, Tong and Chen, Enhong},
title = {{DepFormer}: A Unified Framework with Bimodal Collaborative Transformer for Depression Detection},
year = {2025},
isbn = {9798400720352},
publisher = {Association for Computing Machinery},
address = {New York, NY, USA},
url = {https://doi.org/10.1145/3746027.3762062},
doi = {10.1145/3746027.3762062},
booktitle = {Proceedings of the 33rd {ACM} International Conference on Multimedia},
pages = {13930–13936},
numpages = {7},
location = {Dublin, Ireland},
series = {MM '25}
}

@incollection{10943944,
  author={Kumar, Puneet and Misra, Shreshtha and Shao, Zhuhong and Zhu, Bin and Raman, Balasubramanian and Li, Xiaobai},
  booktitle={2025 {IEEE}/{CVF} Winter Conference on Applications of Computer Vision ({WACV})}, 
  title={Multimodal Interpretable Depression Analysis Using Visual, Physiological, Audio and Textual Data},
  year={2025},
  volume={},
  number={},
  pages={5305-5315},
  doi={10.1109/WACV61041.2025.00518}}

@ARTICLE{10251764,
  author={Yang, Biao and Wang, Peng and Cao, Miaomiao and Zhu, Xianlin and Wang, Suhong and Ni, Rongrong and Yang, Changchun},
  journal={IEEE Transactions on Computational Social Systems}, 
  title={Uncertainty-Aware Label Contrastive Distribution Learning for Automatic Depression Detection},
  year={2024},
  volume={11},
  number={2},
  pages={2979-2989},
  doi={10.1109/TCSS.2023.3311013}}

@incollection{NEURIPS2020_aab08546,
 author = {Amini, Alexander and Schwarting, Wilko and Soleimany, Ava and Rus, Daniela},
 booktitle = {Advances in Neural Information Processing Systems},
 editor = {H. Larochelle and M. Ranzato and R. Hadsell and M.F. Balcan and H. Lin},
 pages = {14927--14937},
 publisher = {Curran Associates, Inc.},
 title = {Deep Evidential Regression},
 url = {https://proceedings.neurips.cc/paper_files/paper/2020/file/aab085461de182608ee9f607f3f7d18f-Paper.pdf},
 volume = {33},
 year = {2020}
}

@article{ma2021trustworthy,
  title={Trustworthy multimodal regression with mixture of normal-inverse gamma distributions},
  author={Ma, Huan and Han, Zongbo and Zhang, Changqing and Fu, Huazhu and Zhou, Joey Tianyi and Hu, Qinghua},
  journal={Advances in Neural Information Processing Systems},
  volume={34},
  pages={6881--6893},
  year={2021}
}

@article{Wu_Shi_Dong_Zheng_Wei_2024, title={The Evidence Contraction Issue in Deep Evidential Regression: Discussion and Solution}, volume={38}, url={https://ojs.aaai.org/index.php/AAAI/article/view/30172}, DOI={10.1609/aaai.v38i19.30172}, number={19}, journal={Proceedings of the AAAI Conference on Artificial Intelligence}, author={Wu, Yuefei and Shi, Bin and Dong, Bo and Zheng, Qinghua and Wei, Hua}, year={2024}, month={3}, pages={21726-21734} }

@misc{silero2024vad,
  author       = {{Silero Team}},
  title        = {{Silero} {VAD}: pre-trained enterprise-grade Voice Activity Detector ({VAD}), Number Detector and Language Classifier},
  year         = {2024},
  publisher    = {GitHub},
  url = {https://github.com/snakers4/silero-vad}
}

@ARTICLE{10572478,
  author={Xu, Jiaqi and Gunes, Hatice and Kusumam, Keerthy and Valstar, Michel and Song, Siyang},
  journal={IEEE Transactions on Affective Computing}, 
  title={Two-Stage Temporal Modelling Framework for Video-Based Depression Recognition Using Graph Representation},
  year={2025},
  volume={16},
  number={1},
  pages={161-178},
  doi={10.1109/TAFFC.2024.3415770}}

@ARTICLE{10851289,
  author={Niu, Mingyue and Wang, Xu and Gong, Jibing and Liu, Bin and Tao, Jianhua and Schuller, Björn W.},
  journal={IEEE Transactions on Circuits and Systems for Video Technology}, 
  title={Depression Scale Dictionary Decomposition Framework for Multimodal Automatic Depression Level Prediction},
  year={2025},
  volume={35},
  number={6},
  pages={6195-6210},
  doi={10.1109/TCSVT.2025.3533480}}

@ARTICLE{9786615,
  author={Uddin, Md Azher and Joolee, Joolekha Bibi and Sohn, Kyung-Ah},
  journal={IEEE Transactions on Affective Computing}, 
  title={Deep Multi-Modal Network Based Automated Depression Severity Estimation},
  year={2023},
  volume={14},
  number={3},
  pages={2153-2167},
  doi={10.1109/TAFFC.2022.3179478}}

@article{FU2025129106,
title = {Facial action units guided graph representation learning for multimodal depression detection},
journal = {Neurocomputing},
volume = {619},
pages = {129106},
year = {2025},
issn = {0925-2312},
doi = {10.1016/j.neucom.2024.129106},
url = {https://www.sciencedirect.com/science/article/pii/S0925231224018770},
author = {Changzeng Fu and Fengkui Qian and Yikai Su and Kaifeng Su and Siyang Song and Mingyue Niu and Jiaqi Shi and Zhigang Liu and Chaoran Liu and Carlos Toshinori Ishi and Hiroshi Ishiguro},
}

@article{MSN,
  title={A deep multiscale spatiotemporal network for assessing depression from facial dynamics},
  author={De Melo, Wheidima Carneiro and Granger, Eric and Hadid, Abdenour},
  journal={IEEE Transactions on Affective Computing},
  volume={13},
  number={3},
  pages={1581--1592},
  year={2020},
  publisher={IEEE}
}

@ARTICLE{MAFF,
  author={Niu, Mingyue and Tao, Jianhua and Liu, Bin and Huang, Jian and Lian, Zheng},
  journal={IEEE Transactions on Affective Computing}, 
  title={Multimodal Spatiotemporal Representation for Automatic Depression Level Detection},
  year={2023},
  volume={14},
  number={1},
  pages={294-307},
  doi={10.1109/TAFFC.2020.3031345}}

@article{AVA-DepressNet,
  title={Integrating deep facial priors into landmarks for privacy preserving multimodal depression recognition},
  author={Pan, Yuchen and Shang, Yuanyuan and Shao, Zhuhong and Liu, Tie and Guo, Guodong and Ding, Hui},
  journal={IEEE Transactions on Affective Computing},
  volume={15},
  number={3},
  pages={828--836},
  year={2023},
  publisher={IEEE}
}

@ARTICLE{MTDAN,
  author={Zhang, Shiqing and Zhang, Xingnan and Zhao, Xiaoming and Fang, Jiangxiong and Niu, Mingyue and Zhao, Ziping and Yu, Jun and Tian, Qi},
  journal={IEEE Transactions on Affective Computing}, 
  title={{MTDAN}: A Lightweight Multi-Scale Temporal Difference Attention Networks for Automated Video Depression Detection},
  year={2024},
  volume={15},
  number={3},
  pages={1078-1089},
  doi={10.1109/TAFFC.2023.3312263}}

@article{DepressNet,
  title={Visually interpretable representation learning for depression recognition from facial images},
  author={Zhou, Xiuzhuang and Jin, Kai and Shang, Yuanyuan and Guo, Guodong},
  journal={IEEE Transactions on Affective Computing},
  volume={11},
  number={3},
  pages={542--552},
  year={2018},
  publisher={IEEE}
}

@article{DJ-LDML,
  title={Facial depression recognition by deep joint label distribution and metric learning},
  author={Zhou, Xiuzhuang and Wei, Zeqiang and Xu, Min and Qu, Shan and Guo, Guodong},
  journal={IEEE Transactions on Affective Computing},
  volume={13},
  number={3},
  pages={1605--1618},
  year={2020},
  publisher={IEEE}
}

@article{AD-DCNN,
  title={Automated depression diagnosis based on deep networks to encode facial appearance and dynamics},
  author={Zhu, Yu and Shang, Yuanyuan and Shao, Zhuhong and Guo, Guodong},
  journal={IEEE Transactions on Affective Computing},
  volume={9},
  number={4},
  pages={578--584},
  year={2017},
  publisher={IEEE}
}

@article{LIU2025105359,
title = {Attention-Guided Bi-direction Temporal-aware Network for speech-based depression recognition},
journal = {Digital Signal Processing},
volume = {166},
pages = {105359},
year = {2025},
issn = {1051-2004},
doi = {10.1016/j.dsp.2025.105359},
url = {https://www.sciencedirect.com/science/article/pii/S1051200425003811},
author = {Jingyi Liu and Yuanyuan Shang and Mengyuan Yang and Zhuhong Shao and Hui Ding and Tie Liu},
}

@article{PAN2024105704,
title = {Multi-feature deep supervised voiceprint adversarial network for depression recognition from speech},
journal = {Biomedical Signal Processing and Control},
volume = {89},
pages = {105704},
year = {2024},
issn = {1746-8094},
doi = {10.1016/j.bspc.2023.105704},
url = {https://www.sciencedirect.com/science/article/pii/S1746809423011370},
author = {Yuchen Pan and Yuanyuan Shang and Wei Wang and Zhuhong Shao and Zhuojin Han and Tie Liu and Guodong Guo and Hui Ding},
}

@article{AFN,
  title={Multitask representation learning for multimodal estimation of depression level},
  author={Qureshi, Syed Arbaaz and Saha, Sriparna and Hasanuzzaman, Mohammed and Dias, Ga{\"e}l},
  journal={IEEE Intelligent Systems},
  volume={34},
  number={5},
  pages={45--52},
  year={2019},
  publisher={IEEE}
}

@incollection{HOG-PCA,
  title={A random forest regression method with selected-text feature for depression assessment},
  author={Sun, Bo and Zhang, Yinghui and He, Jun and Yu, Lejun and Xu, Qihua and Li, Dongliang and Wang, Zhaoying},
  booktitle={Proceedings of the 7th annual workshop on Audio/Visual emotion challenge},
  pages={61--68},
  year={2017}
}

@article{heDeepLearningDepression2022,
  title = {Deep Learning for Depression Recognition with Audiovisual Cues: A Review},
  shorttitle = {Deep Learning for Depression Recognition with Audiovisual Cues},
  author = {He, Lang and Niu, Mingyue and Tiwari, Prayag and Marttinen, Pekka and Su, Rui and Jiang, Jiewei and Guo, Chenguang and Wang, Hongyu and Ding, Songtao and Wang, Zhongmin and Pan, Xiaoying and Dang, Wei},
  year = 2022,
  month = apr,
  journal = {Information Fusion},
  volume = {80},
  pages = {56--86},
  issn = {15662535},
  doi = {10.1016/j.inffus.2021.10.012},
  urldate = {2025-12-09},
  langid = {english}
}

@ARTICLE{11217233,
  author={Gao, Junyu and Chen, Mengyuan and Xiang, Liangyu and Xu, Changsheng},
  journal={IEEE Transactions on Pattern Analysis and Machine Intelligence}, 
  title={A Comprehensive Survey on Evidential Deep Learning and its Applications},
  year={2026},
  volume={48},
  number={3},
  pages={2118-2138},
  doi={10.1109/TPAMI.2025.3625258}}

@ARTICLE{10935667,
  author={Li, Yonghong and Wei, Zeqiang and Guo, Guodong and Zhou, Xiuzhuang},
  journal={IEEE Transactions on Affective Computing}, 
  title={{MemRank}: Memory-Augmented Similarity Ranking for Video-Based Depression Severity Estimation},
  year={2025},
  volume={16},
  number={3},
  pages={2062-2073},
  doi={10.1109/TAFFC.2025.3553090}}

@ARTICLE{11488689,
  author={Wang, Xuzhi and Wu, Xinran and Zhao, Ziping and Tao, Jianhua and Schuller, Bj{\"o}rn W.},
  journal={IEEE Transactions on Affective Computing}, 
  title={{MA-DLE}: Speech-based Automatic Depression Level Estimation via Memory Augmentation},
  year={2026},
  volume={},
  number={},
  pages={1-13},
  doi={10.1109/TAFFC.2026.3685963}}

@ARTICLE{10932863,
  author={Wang, Yaowei and Lin, Zulong and Teng, Yan and Cheng, Yuqi and Jiang, Hongyan and Yang, Yun},
  journal={IEEE Transactions on Computational Social Systems}, 
  title={{SIMMA}: Multimodal Automatic Depression Detection via Spatiotemporal Ensemble and Cross-Modal Alignment},
  year={2025},
  volume={12},
  number={5},
  pages={3548-3564},
  doi={10.1109/TCSS.2025.3542986}}

@ARTICLE{11153053,
  author={Liu, Chengguang and Wang, Shanmin and Liu, Qingshan and Wang, Yang and Wang, Fei},
  journal={IEEE Transactions on Affective Computing}, 
  title={Mitigating Symptom Heterogeneity in Multimodal Depression Estimation via Level Separation and Deviation Regression},
  year={2026},
  volume={17},
  number={1},
  pages={107-118},
  doi={10.1109/TAFFC.2025.3606949}}

@article{yang2026disnet,
  author  = {Yang, Wenju and Cao, Peng and Wang, Yang and Yang, Jie and Wang, Fei},
  title   = {{DisNet}: Learning interpretable depression representations in speech},
  journal = {Neural Networks},
  year    = {2026},
  volume  = {200},
  pages     = {108820},
  doi     = {10.1016/j.neunet.2026.108820}
}

@article{he2022noisyannotations,
  author  = {He, Lang and Tiwari, Prayag and Lv, Chonghua and Wu, Wen Shuai and Guo, Liyong},
  title   = {Reducing noisy annotations for depression estimation from facial images},
  journal = {Neural Networks},
  year    = {2022},
  volume  = {153},
  pages   = {120--129},
  doi     = {10.1016/j.neunet.2022.05.025}
}

@article{ye2025bpen,
  author  = {Ye, Kai and Tang, Haoteng and Dai, Siyuan and Fortel, Igor and Thompson, Paul M. and Mackin, R. Scott and Leow, Alex and Huang, Heng and Zhan, Liang},
  title   = {{BPEN}: Brain Posterior Evidential Network for trustworthy brain imaging analysis},
  journal = {Neural Networks},
  year    = {2025},
  volume  = {183},
  pages     = {106943},
  doi     = {10.1016/j.neunet.2024.106943}
}

@article{yang2023filterbanks,
  author  = {Yang, Wenju and Liu, Jiankang and Cao, Peng and Zhu, Rongxin and Wang, Yang and Liu, Jian K. and Wang, Fei and Zhang, Xizhe},
  title   = {Attention guided learnable time-domain filterbanks for speech depression detection},
  journal = {Neural Networks},
  year    = {2023},
  volume  = {165},
  pages   = {135--149},
  doi     = {10.1016/j.neunet.2023.05.041}
}

@article{sadok2024mdvae,
  author  = {Sadok, Samir and Leglaive, Simon and Girin, Laurent and Alameda-Pineda, Xavier and S{\'e}guier, Renaud},
  title   = {A multimodal dynamical variational autoencoder for audiovisual speech representation learning},
  journal = {Neural Networks},
  year    = {2024},
  volume  = {172},
  pages     = {106120},
  doi     = {10.1016/j.neunet.2024.106120}
}

@article{nilsen2022delta,
  author  = {Nilsen, Geir K. and Munthe-Kaas, Antonella Z. and Skaug, Hans J. and Brun, Morten},
  title   = {Epistemic uncertainty quantification in deep learning classification by the {Delta} method},
  journal = {Neural Networks},
  year    = {2022},
  volume  = {145},
  pages   = {164--176},
  doi     = {10.1016/j.neunet.2021.10.014}
}

@inproceedings{zellinger2017cmd,
  author = {Zellinger, Werner and Grubinger, Thomas and Lughofer, Edwin and Natschl{\"a}ger, Thomas and Saminger-Platz, Susanne},
  title = {Central Moment Discrepancy ({CMD}) for Domain-Invariant Representation Learning},
  booktitle = {International Conference on Learning Representations},
  year = {2017},
  url = {https://arxiv.org/abs/1702.08811}
}

\end{document}